\documentclass[11pt,a4paper]{article}
\usepackage[hyperref]{emnlp-ijcnlp-2019}
\usepackage{times}
\usepackage{latexsym}
\usepackage{amsmath}
\usepackage{amssymb}
\usepackage{url}
\usepackage{graphicx}
\usepackage{array}

\usepackage{xspace}
\aclfinalcopy % Uncomment this line for the final submission

\newcommand\RE{$RE$\xspace}
\renewcommand\OE{$OE$\xspace}
\newcommand\REs{$RE$s\xspace}
\newcommand\OEs{$OE$s\xspace}
\newcommand{\commentout}[1]{}

\newenvironment{para_packed_item}{
\begin{changemargin}{-1.75em}{0em} 
\begin{itemize}
  \setlength{\itemsep}{1pt}
  \setlength{\parskip}{0pt}
  \setlength{\parsep}{0pt}
   \setlength{\itemindent}{0.9em}
}{\end{itemize}
\end{changemargin}
}

\newenvironment{para_packed_enum}{
\begin{changemargin}{-1.75em}{0em} 
\begin{enumerate}
  \setlength{\itemsep}{1pt}
  \setlength{\parskip}{0pt}
  \setlength{\parsep}{0pt}
   \setlength{\itemindent}{0.9em}
}{\end{enumerate}
\end{changemargin}
}

\title{Keep Calm and Switch On! Preserving Sentiment and Fluency in Semantic Text Exchange}

\renewcommand*{\thefootnote}{\fnsymbol{footnote}}

\author{Steven Y. Feng\footnotemark[1]\\ \\\And
  Aaron W. Li\footnotemark[1]\\ \smallskip
  \\ 
  David R. Cheriton School of Computer Science\\
  University of Waterloo\\
  Waterloo, Ontario, Canada\\
  \{sy2feng, w89li, jhoey\}@uwaterloo.ca\\\And
    Jesse Hoey \\
  \\
  }

\date{}

\begin{document}
\maketitle
\footnotetext[1]{Authors contributed equally}

\renewcommand*{\thefootnote}{\arabic{footnote}}
\begin{abstract}
  %NLP researchers have long been utilizing token and sentence embeddings as means of representing text. Recent advancements have opened up the possibility of
  %Recent advancements further demonstrated effective capturing of semantic roles, sentiment, and less biased associations.
  %Encoding sentence level embeddings along with the effective separation of different parts within sentences opens up possibilities. In particular, it enables the task of
  %semantically adjusting a ground truth sentence written by a human or machine. 
  %The ability to perform this targeted task has many potential applications, and we briefly mention a few here. Firstly, the model is able to modify human-written sentences. Secondly, it can be used as a tool to correct the content of machine-generated text. 
  In this paper, we present a novel method for measurably adjusting the semantics of text while preserving its sentiment and fluency, a task we call \textit{semantic text exchange}. This is useful for text data augmentation and the semantic correction of text generated by chatbots and virtual assistants. We introduce a pipeline called \textit{SMERTI} that combines entity replacement, similarity masking, and text infilling. We measure our pipeline's success by its \textit{Semantic Text Exchange Score (STES)}: the ability to preserve the original text's sentiment and fluency while adjusting semantic content. We propose to use \textit{masking (replacement) rate threshold} as an adjustable parameter to control the amount of semantic change in the text. Our experiments demonstrate that SMERTI can outperform baseline models on Yelp reviews, Amazon reviews, and news headlines.
\end{abstract}

\section{Introduction}
%With the rising popularity of Natural Language Processing (NLP) has come an increased demand for text data. Combined with the growth of social media, chatbots, and virtual assistants, one task that has surfaced with increasing importance is the ability to independently adjust text semantics (e.g. its content) and style (e.g. its sentiment).
%generating text with style, sentiment, and more recently distinct personas and personalities, 

There has been significant research on style transfer, with the goal of changing the style of text while preserving its semantic content. The alternative where semantics are adjusted while keeping style intact, which we call {\em semantic text exchange (STE)}, has not been investigated to the best of our knowledge.
%We define this new task as semantic text exchange, and it can be used to modify text that is either human-written or machine-generated. To the best of our knowledge, there has been no proposed model designed specifically for the task of semantic text exchange, especially in a focused and rigorous manner.To tackle this, w
Consider the following example, where the replacement entity defines the new semantic context:
%We call the text component containing the semantic content we desire the {\bf Replacement Entity (RE)}.
\mbox{}\vspace{1em}
\framebox{
  \begin{minipage}{0.95\columnwidth}
\noindent \underline{Original Text}: {\em It is {\bf\em sunny} outside! Ugh, that means I must {\bf\em wear sunscreen}. I hate being {\bf\em sweaty} and {\bf\em sticky all over.}}\\
\underline{Replacement Entity}: weather = {\bf{\em rainy}}\\
\underline{Desired Text}: \emph{It is {\bf\em rainy} outside! Ugh, that means I must {\bf\em bring an umbrella}. I hate being {\bf\em wet} and {{\bf\em having to carry it around.}}}
\end{minipage}
}\vspace{1em}
%In this case, we can see that
The weather within the original text is sunny, whereas the actual weather may be rainy. %The goal is to exchange the original text to the desired text.
Not only is the word \textit{sunny} replaced with \textit{rainy}, but the rest of the text's content is changed while preserving its negative sentiment and fluency. %These changes are bolded.
%This is because it can be either a word or a phrase, and represents an overall entity. It is also the main entity that we wish to insert into the original text.

With the rise of natural language processing (NLP) has come an increased demand for massive amounts of text data. Manually collecting and scraping data requires a significant amount of time and effort, and data augmentation techniques for NLP are limited compared to fields such as computer vision. STE can be used for text data augmentation by producing various modifications of a piece of text that differ in semantic content.

Another use of STE is in building emotionally aligned chatbots and virtual assistants. This is useful for reasons such as marketing, overall enjoyment of interaction, and mental health therapy. However, due to limited data with emotional content in specific semantic contexts, the generated text may contain incorrect semantic content. STE can adjust text semantics (e.g. to align with reality or a specific task) while preserving emotions.

One specific example is the development of virtual assistants with adjustable socio-emotional personalities in the effort to construct assistive technologies for persons with cognitive disabilities. Adjusting the emotional delivery of text in subtle ways can have a strong effect on the adoption of the technologies~\cite{robillard2018ethical}. It is challenging to transfer style this subtly due to lack of datasets on specific topics with consistent emotions. Instead, large datasets of emotionally consistent interactions not confined to specific topics exist. Hence, it is effective to generate text with a particular emotion and then adjust its semantics.
%% to be put in the final version~\cite{RobillardHoey2018,Robillard-EthicalAdoption-2018,Konig2017}.

We propose a pipeline called SMERTI (pronounced `\textit{smarty}') for STE.\footnote{Code for SMERTI (including Google Colab links) can be found at \href{https://github.com/styfeng/SMERTI}{https://github.com/styfeng/SMERTI}} Combining entity replacement (ER), similarity masking (SM), and text infilling (TI), SMERTI can modify the semantic content of text. We define a metric called the {\em Semantic Text Exchange Score (STES)} that evaluates the overall ability of a model to perform STE, and an adjustable parameter \textit{masking (replacement) rate threshold (MRT/RRT)} that can be used to control the amount of semantic change.

%We train two versions of our pipeline: SMERTI-RNN and SMERTI-Transformer. The former's text infilling module is trained on an RNN model~\cite{sutskever2014sequence}, while the latter's text infilling module is trained on a transformer model~\cite{vaswani2017attention}.
We evaluate on three datasets: Yelp and Amazon reviews~\cite{he2016ups}, and Kaggle news headlines~\cite{Misra2018}. %The first two datasets are specifically for social media reviews and carry significant sentiment, while the third contains more neutral and informative text.
We implement three baseline models for comparison: Noun WordNet Semantic Text Exchange Model (NWN-STEM), General WordNet Semantic Text Exchange Model (GWN-STEM), and Word2Vec Semantic Text Exchange Model (W2V-STEM).

We illustrate the STE performance of two SMERTI variations on the datasets, demonstrating outperformance of the baselines and pipeline stability. We also run a human evaluation supporting our results. We analyze the results in detail and investigate relationships between the semantic change, fluency, sentiment, and MRT/RRT. %by model, dataset, part-of-speech, and amount of text masked and/or replaced.
%Through our experiments, we discover that masking rate plays a strong role in the ability of the pipeline to preserve fluency and sentiment. We define an adjustable parameter called the masking rate threshold, which represents an upper-bound on the percentage of the original text that can be modified. A higher masking rate threshold gives SMERTI the ability to alter more of the original text, while a lower threshold lessens its ability to modify the original text. We find that higher thresholds lead to increased semantic differentiation between input and output text, but lower fluency and sentiment preservation. On the other hand, lower thresholds restrict semantic differentiation while keeping fluency high and preserving sentiment.
%Overall,
Our major contributions can be summarized as:
\begin{para_packed_item}
\item We define a new task called {\em semantic text exchange (STE)} with increasing importance in NLP applications that modifies text semantics while preserving other aspects such as sentiment.
\item We propose a pipeline {\em SMERTI} capable of multi-word entity replacement and text infilling, and demonstrate its outperformance of baselines.
\item We define an evaluation metric for overall performance on semantic text exchange called the {\em Semantic Text Exchange Score (STES)}. %that takes into account various aspects of the resulting text indicative of strong performance. 
%\item We define an adjustable parameter \textit{masking (replacement) rate threshold} that controls the amount of semantic change in the text
\end{para_packed_item}

\section{Related Work}
\subsection{Word and Sentence-level Embeddings}
Word2Vec~\cite{mikolov2013efficient, mikolov2013distributed} allows for analogy representation through vector arithmetic. We implement a baseline (W2V-STEM) using this technique. The Universal Sentence Encoder (USE)~\cite{cer-etal-2018-universal} encodes sentences and is trained on a variety of web sources and the Stanford Natural Language Inference corpus~\cite{bowman-etal-2015-large}. Flair embeddings~\cite{akbik2018contextual} are based on architectures such as BERT~\cite{devlin-etal-2019-bert}. We use USE for SMERTI as it is designed for transfer learning and shows higher performance on textual similarity tasks compared to other models \cite{perone2018evaluation}.

\subsection{Text Infilling}
Text infilling is the task of filling in missing parts of sentences called masks. MaskGAN~\cite{fedus2018maskgan} is restricted to a single word per mask token, while SMERTI is capable of variable length infilling for more flexible output. \citet{zhu2019text} uses a transformer-based architecture. They fill in random masks, while SMERTI fills in masks guided by semantic similarity, resulting in more natural infilling and fulfillment of the STE task.

\subsection{Style and Sentiment Transfer}
%Style and sentiment transfer basically tries to achieve the reverse of the semantic text exchange task SMERTI is designed for:
%Their goal is to change the sentiment or style of text while keeping semantics and content intact~\cite{huang2017arbitrary, Logeswaran2018}.
%For example, converting a positive sentence to a negative one, or converting a sentence that has a professional tone to one that is more informal. Some
Notable works in style/sentiment transfer include~\cite{shen2017style, fu2018style, li-etal-2018-delete, xu-etal-2018-unpaired}. They attempt to learn latent representations of various text aspects such as its context and attributes, or separate style from content and encode them into hidden representations. They then use an RNN decoder to generate a new sentence given a targeted sentiment attribute.
%Unlike these works, we investigate the complementary task of semantic text exchange given a replacement entity. Our overall pipeline differs significantly as we use sentence-level embeddings to calculate semantic similarities and determine which words should be replaced, followed by text infilling by training masked language models. 

\subsection{Review Generation}
    \citet{hovy2016enemy} generates fake reviews from scratch using language models.~\cite{lipton2015capturing, dong2017learning, juuti2018stay} generate reviews from scratch given auxiliary information (e.g. the item category and star rating). ~\citet{yao2017automated} generates reviews using RNNs with two components: generation from scratch and review customization (Algorithm 2 in ~\citet{yao2017automated}). They define review customization as modifying the generated review to fit a new topic or context, such as from a Japanese restaurant to an Italian one. They condition on a keyword identifying the desired context, and replace similar nouns with others using WordNet ~\cite{miller1995wordnet}. They require a \textit{``reference dataset"} (required to be \textit{``on topic"}; easy enough for restaurant reviews, but less so for arbitrary conversational agents). As noted by~\citet{juuti2018stay}, the method of~\citet{yao2017automated} may also replace words independently of context. We implement their review customization algorithm (NWN-STEM) and a modified version (GWN-STEM) as baseline models.

\section{SMERTI}
%In this section, we introduce our pipeline SMERTI. We first give an overview of our pipeline and describe the overall architecture. Then, we discuss the major components of our pipeline in more detail.
\subsection{Overview}
\begin{figure}
\begin{tabular}{c}
\includegraphics[width=0.45\textwidth]{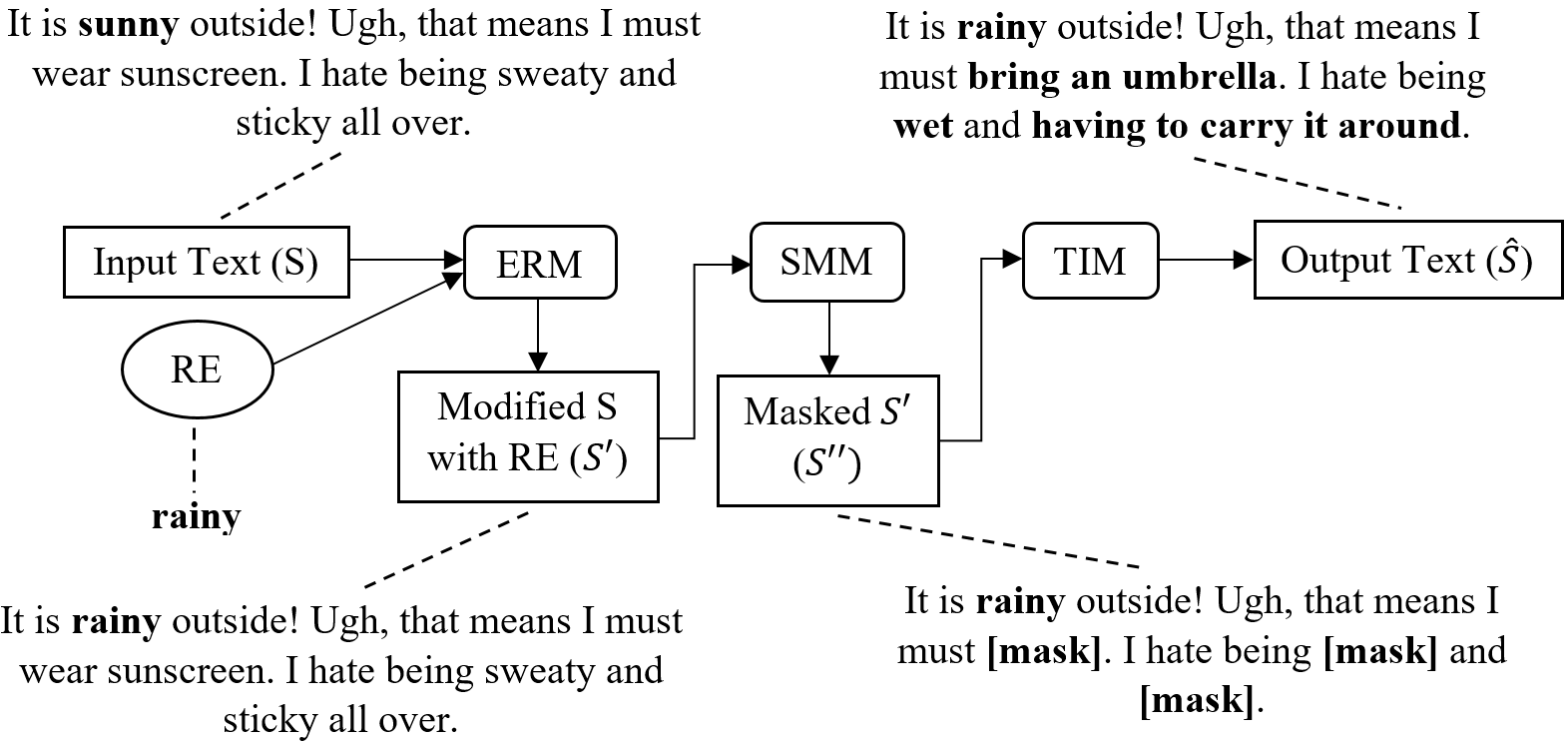}
\end{tabular}
  \caption{\label{fig:overall} Overall architecture and example, showing the three modules: Entity Replacement (ERM), Similarity Masking (SMM), and Text Infilling (TIM)}
\end{figure}

The task is to transform a corpus $C$ of lines of text $S_i$ and associated replacement entities $RE_i:C = \{(S_1,RE_1),(S_2,RE_2),\ldots, (S_n, RE_n)\}$ to a modified corpus $\hat{C} = \{\hat{S}_1,\hat{S}_2,\ldots,\hat{S}_n\}$, where $\hat{S}_i$ are the original text lines $S_i$ replaced with $RE_i$ and overall semantics adjusted.
SMERTI consists of the following modules, shown in Figure~\ref{fig:overall}: %major steps and modules:
\begin{para_packed_enum}
\item Entity Replacement Module (ERM): Identify which word(s) within the original text are best replaced with the \RE, which we call the Original Entity (\OE). We replace \OE in $S$ with \RE. We call this modified text $S'$.
\item Similarity Masking Module (SMM): Identify words/phrases in $S'$ similar to \OE and replace them with a {\em [mask]}. Group adjacent {\em [mask]}s into a single one so we can fill a variable length of text into each. We call this masked text $S''$.
\item 	Text Infilling Module (TIM): Fill in {\em [mask]} tokens with text that better suits the \RE. This will modify semantics in the rest of the text. This final output text is called $\hat{S}$.
\end{para_packed_enum}
%Our overall architecture along with an example is illustrated in Figure~\ref{fig:overall}.
%%JH: We will shrink these figures later. (a) and (b) can easily be combined into a single, more compact figure ADDRESSED
\subsection{Entity Replacement Module (ERM)}
For entity replacement, we use a combination of the Universal Sentence Encoder~\cite{cer-etal-2018-universal} and Stanford Parser~\cite{chen2014fast}.

\subsubsection*{Stanford Parser}
The Stanford Parser is a constituency parser that determines the grammatical structure of sentences, including phrases and part-of-speech (POS) labelling. By feeding our \RE through the parser, we are able to determine its parse-tree. Iterating through the parse-tree and its sub-trees, we can obtain a list of constituent tags for the \RE. We then feed our input text $S$ through the parser, and through a similar process, we can obtain a list of leaves (where leaves under a single label are concatenated) that are equal or similar to any of the \RE constituent tags. This generates a list of entities having the same (or similar) grammatical structure as the \RE, and are likely candidates for the \OE. We then feed these entities along with the \RE into the Universal Sentence Encoder (USE). %in Section~\ref{sec:USE}. 

%One advantage of using the Stanford Parser is that we are able to handle \REs and \OEs which are multiple words long. In particular, we are able to perform word-to-word, phrase-to-word, word-to-phrase, and phrase-to-phrase replacement. The Stanford Parser can identify both words and phrases which are potentially suitable \OEs to be replaced. Another advantage of using the Stanford Parser is that we are able to determine the POS of the \RE. As such, we are able to replace a variety of words ranging from nouns to verbs, adjectives, and so forth. Our module is designed to handle \REs of various lengths and POS.

\subsubsection*{Universal Sentence Encoder (USE)}
\label{sec:USE}
The USE is a sentence-level embedding model
%It is designed for encoding sentences into embedding vectors with the purpose of targeting transfer learning to other NLP tasks. One of these tasks is semantic relatedness and textual similarity, and USE performs extremely well compared to other embedding models as demonstrated in \citet{perone2018evaluation}.
that comes with a deep averaging network (DAN) and transformer model~\cite{cer-etal-2018-universal}. We choose the transformer model as these embeddings take context into account, and the exact same word/phrase will have a different embedding depending on its context and surrounding words. 

We compute the semantic similarity between two embeddings $u$ and $v$: $sim(u,v)$, using the angular (cosine) distance, defined as: $\cos(\theta_{u,v}) = (u\cdot v)/(||u|| ||v||)$, such that 
%($angular\_distance(u,v)$. Angular distance is a function of the cosine similarity between U and V ( cos(θ_U ,_V )  ), where θ_U ,_V is the angle between vectors U and V. We then subtract the angular distance from 1 to obtain a final similarity score.
$sim(u,v) = 1-\frac{1}{\pi}arccos(\cos(\theta_{u,v}))$.
Results are in $[0,1]$, with higher values representing greater similarity.

%% add equations
Using USE and the above equation, we can identify words/phrases within the input text $S$ which are most similar to \RE. To assist with this, we use the Stanford Parser as described above to obtain a list of candidate entities. In the rare case that this list is empty, %(meaning it found no words or phrases in S that have the same constituency tags as \RE),
we feed in each word of $S$ into USE, and identify which word is the most similar to \RE. We then replace the most similar entity or word (\OE) with the \RE and generate $S'$.

\begin{figure}
\begin{tabular}{ccc}
\includegraphics[width=0.45\textwidth]{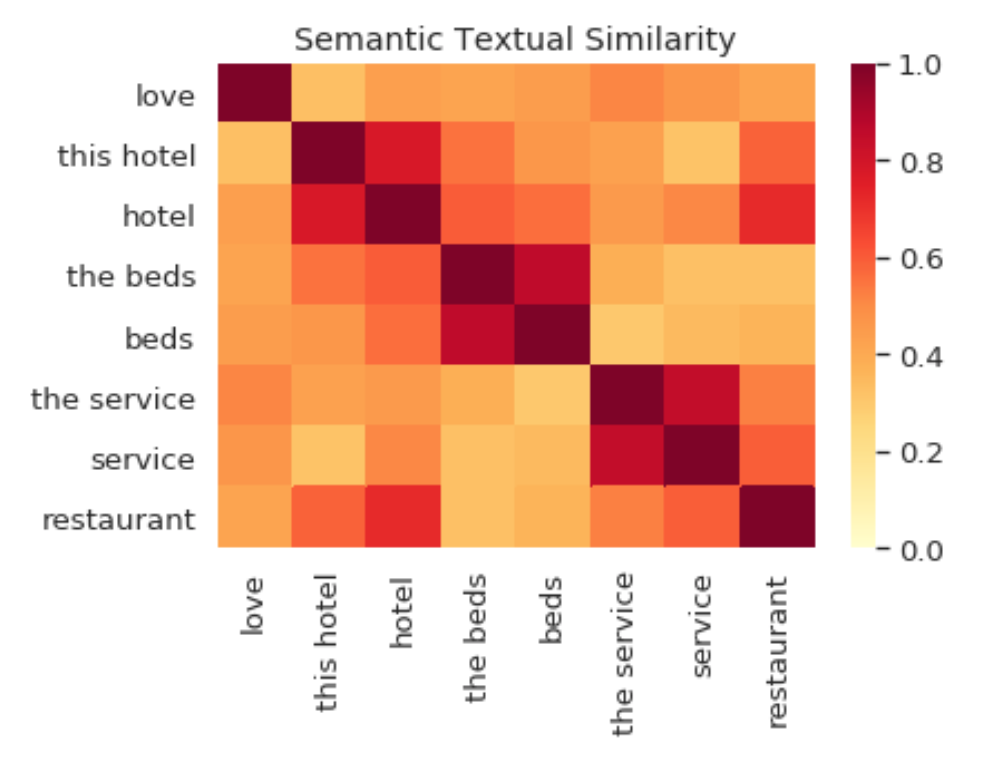} \\
(d) \\
\end{tabular}
  \caption{\label{fig:parser} ERM example with   $S$ = {\em i love this hotel ! the beds are comfortable and the service is great !} and \RE = {\em restaurant} showing (a) Parse tree for \RE; (b) and (c) Parse tree for S; (d) Semantic similarity heat map}
\end{figure}

An example of this entity replacement process is in Figure~\ref{fig:parser}. Two parse-trees are shown: for \RE (a) and $S$ (b) and (c). % as determined by the Stanford Parser.
Figure~\ref{fig:parser}(d) is a semantic similarity heat-map generated from the USE embeddings of the candidate \OEs and \RE, where values are similarity scores in the range $[0,1]$.

\begin{figure*}
\begin{tabular}{ccc}
\includegraphics[width=0.10\textwidth]{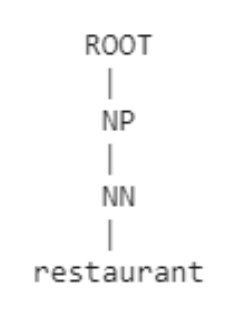} &
\includegraphics[width=0.22\textwidth]{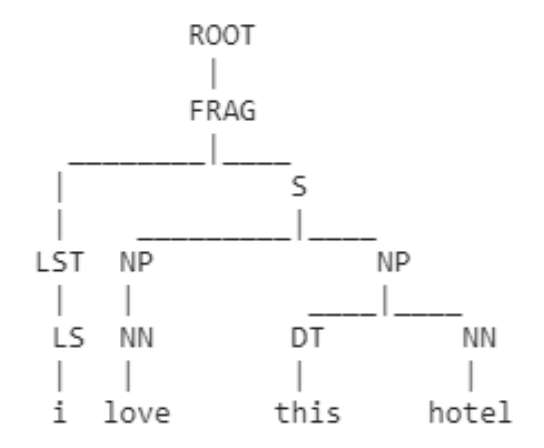} &
\includegraphics[width=0.60\textwidth]{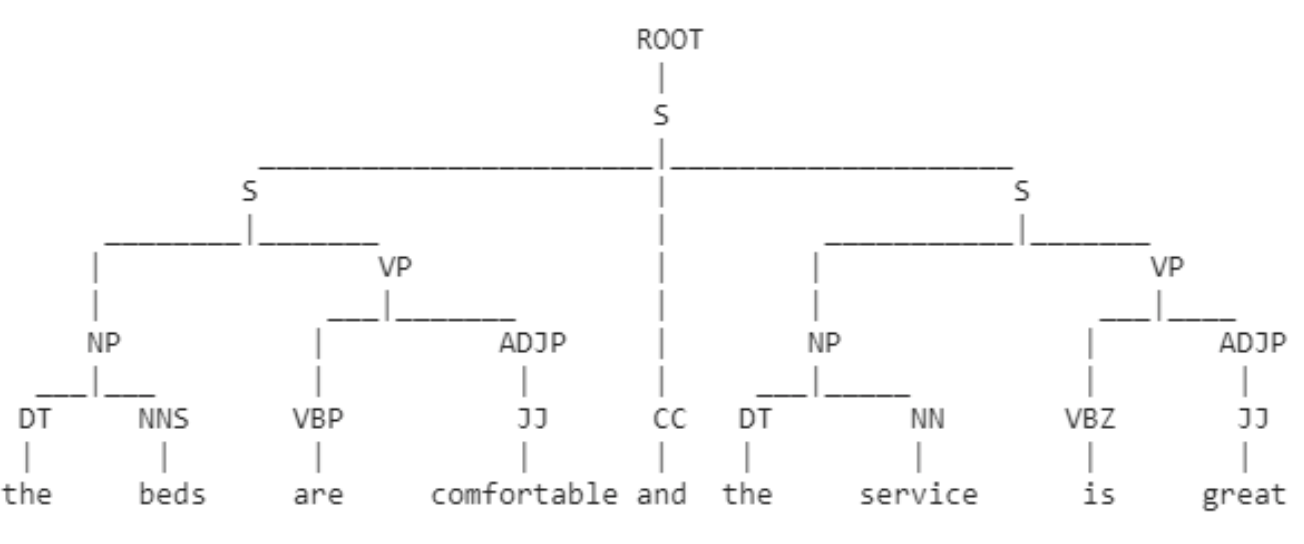} \\
(a) & (b) & (c) \\
\end{tabular}
\end{figure*}

As seen in Figure~\ref{fig:parser}(d), we calculate semantic similarities between \RE and entities within $S$ which have \textit{noun} constituency tags. Looking at the row for our \RE \textit{restaurant}, the most similar entity (excluding itself) is \textit{hotel}. We can then generate:

$S'$ = {\em i love this restaurant ! the beds are comfortable and the service is great !}

%Note that in our actual implementation, we only calculate the semantic similarities between \RE and the candidate entities, rather than every entity in the list against each other to save computation resource and time. The above heat-map is generated for display purposes.

\subsection{Similarity Masking Module (SMM)}
%\subsubsection{Universal Sentence Encoder (USE)}
Next, we mask words similar to \OE to generate $S''$ using USE. We look at semantic similarities between every word in $S$ and \OE, along with semantic similarities between \OE and the candidate entities determined in the previous ERM step to broaden the range of phrases our module can mask. We ignore \RE, \OE, and any entities or phrases containing \OE (for example, \textit{`this hotel'}).

After determining words similar to the \OE (discussed below), we replace each of them with a {\em [mask]} token.
%\subsubsection{Mask Groupings}
Next, we replace {\em [mask]} tokens adjacent to each other with a single {\em [mask]}.
%The reason for doing so is because we wish for our TIM to have the ability to fill in a variable length of text for each {\em [mask]} token. If we do not perform this step, our TIM will only learn to fill each {\em [mask]} token with a single word, resulting in text infillings that are inflexible. This is another advantage of our module compared to the many other text infilling models which only learn to fill in each {\em [mask]} token with a single word.

%\subsubsection{\bf Masking Rate and Similarity Thresholds (MRT and ST)}
%\label{sec:MRTST}
%THIS SECTION IS TOO WORDY, SHOULD BE CUT DOWN
We set a base similarity threshold (ST) that selects a subset of words to mask. We compare the actual fraction of masked words to the masking rate threshold (MRT), as defined by the user, and increase ST in intervals of $0.05$ until the actual masking rate falls below the MRT.\footnote{There are certain cases where two or more outputs for different MRT may be equal. This occurs when a valid ST cannot be found that masks a larger portion of the sentence without going over MRT.}
%How do we determine if a word is ‘similar' enough to \OE to be masked? How can we decide a similarity threshold (ST) such that all words above this value are masked? We approach this problem by firstly determining the maximum proportion of the sentence we would like to mask, and calling this proportion the masking rate threshold (MRT). We set a base ST that we would like to start with, typically a value like 0.1 which will include most words in the sentence. Using this base ST, if we mask greater than MRT proportion of the sentence, our algorithm automatically increments ST by 0.05 each time and attempts to re-mask the sentence until we mask less than MRT proportion. Some default MRT and base ST combinations we use are 0.2 MRT and 0.4 ST, 0.4 MRT and 0.3 ST, 0.6 MRT and 0.2 ST, and 0.8 MRT and 0.1 ST. In fact, these are the four combinations we use for our evaluation later.
Some sample masked outputs ($S''$) using various MRT-ST combinations for the previous example are shown in Table~\ref{tab:tab1} (more examples in Appendix A).

\begin{table}
\begin{tabular}{ccc}
\multicolumn{1}{c}{\includegraphics[width=0.45\textwidth]{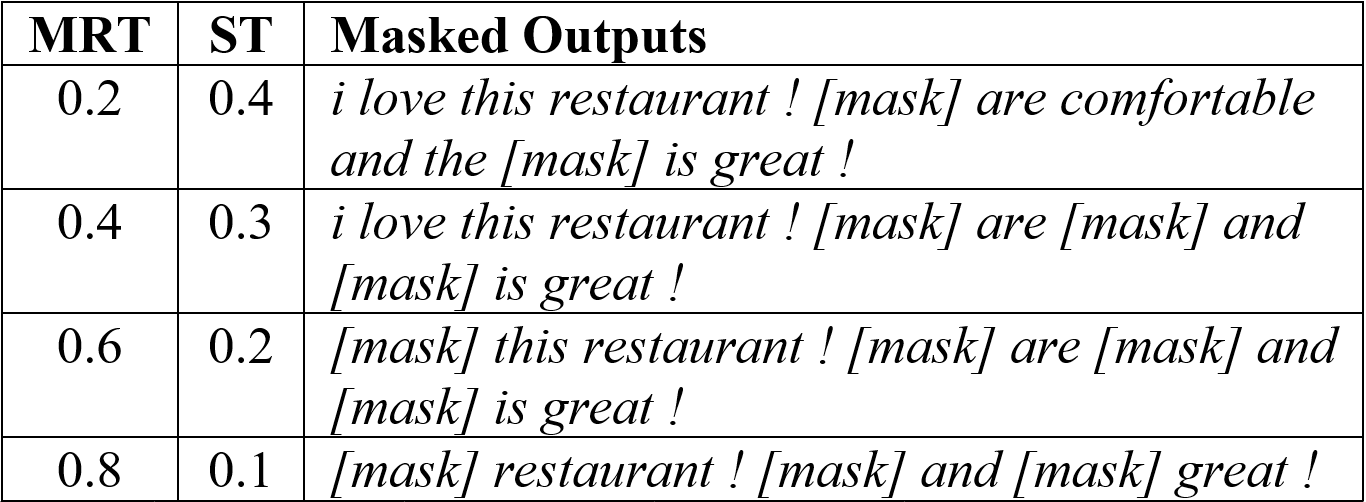}}\\
\end{tabular}
  \caption{\label{tab:tab1} Masked outputs for different masking rate thresholds (MRT) and base similarity thresholds (ST)}
\end{table}

%For example, consider that MRT = 0.4, ST = 0.3 and we mask three words of $S'$. Let's say we now set MRT = 0.6, ST = 0.2, and at this ST value we mask over 60\% of $S'$. We exceed MRT, and hence increase ST to 0.25. Let's say we still mask over 60\% of $S'$, so we increase ST once more to 0.3. In this case, we have the same ST as the previous MRT-ST combination and mask the same three words of $\hat{S}'$, dropping below the current MRT of 0.6, so this will be our final output.

%We believe that the MRT is the most important factor to the user, and the appropriate ST naturally follows. It basically represents the amount the user wishes for the output to differ from the input, particularly in terms of its semantics.
The MRT is similar to the temperature parameter used to control the ``novelty'' of generated text in works such as~\citet{yao2017automated}.
%(again, our main competitor, if we include them the reviewers might question why we didn't benchmark against them).
A high MRT means the user wants to generate text very semantically dissimilar to the original, and may be desired in cases such as creating a lively chatbot or correcting text that is heavily incorrect semantically. A low MRT means the user wants to generate text semantically similar to the original, and may be desired in cases such as text recovery, grammar correction, or correcting a minor semantic error in text. By varying the MRT, various pieces of text that differ semantically in subtle ways can be generated, assisting greatly with text data augmentation. The MRT also affects sentiment and fluency, as we show in Section~\ref{sec:analysis}.

%Further, the amount of masking affects the ability of the text to maintain its sentiment. We can define this as the sentiment preservation (SP), which is the ability of our pipeline to maintain sentiment between input and output. A higher MRT will result in less information being available, and our TIM (discussed in the next section) having to fill in more of the text itself, generally resulting in a lower SP. Hence, if one of the goals is the preservation of some aspect of the text related to style or sentiment, then a lower MRT will likely be better. We show these previous points hold true and discuss them in more detail later in our evaluation and analysis sections.

%Note that numerous more examples of our ERM and SMM outputs on various sample input texts and replacement entities can be found in Appendix A, including more complicated ones.

\subsection{Text Infilling Module (TIM)}
We use two seq2seq models for our TIM: an RNN (recurrent neural network) model~\cite{sutskever2014sequence} (called SMERTI-RNN), and a transformer model (called SMERTI-Transformer). %They are strong seq2seq models designed specifically for text generation, and are suitable for our text infilling purposes where we wish to generate output text from masked input text, as our models learn to fill in [mask] tokens.
%We call our overall pipelines using these two models SMERTI-RNN and SMERTI-Transformer, respectively.

\subsubsection*{Bidirectional RNN with Attention}
%Our first model uses a bidirectional RNN with attention.
We use a bidirectional variant of the GRU~\cite{cho-etal-2014-learning}, and hence two RNNs for the encoder: one reads the input sequence in standard sequential order, and the other is fed this sequence in reverse. The outputs are summed at each time step, giving us the ability to encode information from both past and future context. 

The decoder generates the output in a sequential token-by-token manner. To combat information loss, we implement the attention mechanism~\cite{bahdanau2014neural}. %This gives the decoder the ability to pay attention to specific parts of the input sequence.
We use a Luong attention layer~\cite{luong-etal-2015-effective} which uses global attention, where all the encoder's hidden states are considered, and use the decoder's current time-step hidden state to calculate attention weights. We use the dot score function for attention, where $h_t$ is the current target decoder state and $\bar{h}_s$ is all encoder states: $score(h_t,\bar{h}_s)=h_t^T\bar{h}_s$.

\subsubsection*{Transformer}
Our second model makes use of the transformer architecture, and our implementation replicates \citet{vaswani2017attention}. We use an encoder-decoder structure with a multi-head self-attention token decoder to condition on information from both past and future context. It maps a query and set of key-value pairs to an output. The queries and keys are of dimension $d_k$, and values of dimension $d_v$. To compute the attention, we pack a set of queries, keys, and values into matrices $Q$, $K$, and $V$, respectively. The matrix of outputs is computed as:

\begin{footnotesize}
\begin{equation}
  Attention(Q,K,V)=softmax\left(\frac{QK^T}{\sqrt{d_k}}\right) V
  \label{eqn:attention}
  \end{equation}
\end{footnotesize}

Multi-head attention allows the model to jointly attend to information from different positions. The decoder can make use of both local and global semantic information while filling in each {\em [mask]}.

\section{Experiment}
\label{sec:experiment}
%We evaluated our pipeline on two review datasets: Yelp and Amazon~\cite{he2016ups}, as well as a Kaggle news headlines dataset~\cite{Misra2018} to show our pipeline can achieve strong results on various types of text.
%We will firstly discuss the datasets, the training and evaluation setup and details, and our evaluation metrics. Lastly, we will analyze the results of our experiment.
\subsection{Datasets}
We train our two TIMs on the three datasets. The Amazon dataset~\cite{he2016ups} contains over 83 million user reviews on products, with duplicate reviews removed. The Yelp dataset includes over six million user reviews on businesses. The news headlines dataset from Kaggle contains approximately $200,000$ news headlines from 2012 to 2018 obtained from HuffPost~\cite{Misra2018}.
%This dataset contains headlines corresponding to a variety of categories such as politics, entertainment, and crime. 

We filter the text to obtain reviews and headlines which are English, do not contain hyperlinks and other obvious noise, and are less than 20 words long. We found that many longer than twenty words ramble on and are too verbose for our purposes. Rather than filtering by individual sentences we keep each text in its entirety so SMERTI can learn to generate multiple sentences at once. We preprocess the text by lowercasing and removing rare/duplicate punctuation and space.
%We preprocess the text as well to ensure everything is lowercase, duplicate punctuation and spaces are removed, and so forth.

For Amazon and Yelp, we treat reviews greater than three stars as containing positive sentiment, equal to three stars as neutral, and less than three stars as negative. For each training and testing set, we include an equal number of randomly selected positive and negative reviews, and half as many neutral reviews. This is because neutral reviews only occupy one out of five stars compared to positive and negative which occupy two each.
%Furthermore, neutral reviews carry a lot of noise with them, as many of them carry a relatively high amount of positive or negative sentiment and hence can be classified as either positive or negative. A small proportion of the neutral reviews are actually objectively neutral.
Our dataset statistics can be found in Appendix B.

%%%%%%%%%%%%%%%%%%%%%%%%%5
%% insert table 1 here
%%%%%%%%%%%%%%%%%%%%%%%%%5

%\begin{table*}[t!]
%\centering
%\begin{tabular}{llll}
%  Dataset & Sentiment & Training Set & Testing Set\\
%  \hline
%  Amazon & Positive & 30K  & 5K   \\
%                    & Negative & 30K  & 5K   \\
%                    & Neutral  & 15K  & 2.5K \\
%  Yelp & Positive & 30K  & 5K   \\
%                  & Negative & 30K  & 5K   \\
%                  & Neutral  & 15K  & 2.5K \\
%  News headlines  & & 120K & 20K \\
%\end{tabular}
%\caption{Data statistics.
%  }\label{tab:table1}
%\end{table*}

\subsection{Experiment Details}
%\subsubsection{Masking Data Setup}
To set up our training and testing data for text infilling, we mask the text. We use a tiered masking approach: for each dataset, we randomly mask 15\% of the words in one-third of the lines, 30\% of the words in another one-third, and 45\% in the remaining one-third. %For Amazon and Yelp, we ensure one-third of the reviews for each sentiment is masked with one of these rates. 
These masked texts serve as the inputs, while the original texts serve as the ground-truth. This allows our TIM models to learn relationships between masked words and relationships between masked and unmasked words.

%\subsubsection{SMERTI-RNN}
%The first mask-filling model we train is
The bidirectional RNN decoder fills in blanks one by one, with the objective of minimizing the cross entropy loss between its output and the ground-truth. We use a hidden size of 500, two layers for the encoder and decoder, teacher-forcing ratio of 1.0, learning rate of 0.0001, dropout of 0.1, batch size of 64, and train for up to 40 epochs.

%\subsubsection{SMERTI-Transformer}
For the transformer, we use scaled dot-product attention and the same hyperparameters as ~\citet{vaswani2017attention}. We use the Adam optimizer \cite{kingma2014adam} with $\beta_1 = 0.9, \beta_2 = 0.98$, and $\epsilon = 10^{-9}$. As in~\citet{vaswani2017attention}, we increase the $learning\_rate$ linearly for the first $warmup\_steps$ training steps, and then decrease the $learning\_rate$ proportionally to the inverse square root of the step number. We set $factor=1$ and use $warmup\_steps = 2000$. 
%%%%%%%%%%%%%%%%%%%%%%%%%5
%% insert learning rate equation
%%%%%%%%%%%%%%%%%%%%%%%%%5
We use a batch size of 4096, and we train for up to 40 epochs. 

\subsection{Baseline Models}\label{sec:benchmarks}
We implement three models to benchmark against.\footnote{See Appendix C for more implementation details} First is NWN-STEM (Algorithm 2 from~\citet{yao2017automated}). We use the training sets as the ``reference review sets" to extract similar nouns to the \RE (using {\em MIN\textsubscript{sim}} = 0.1). We then replace nouns in the text similar to the \RE with nouns extracted from the associated reference review set.

Secondly, we modify NWN-STEM to work for verbs and adjectives\footnote{WordNet can only work for single words (and not phrases). Also, it turns out that it cannot work for most adjective \REs, as discussed in Appendix C}, and call this GWN-STEM. From the reference review sets, we extract similar nouns, verbs, and adjectives to the \RE (using  {\em MIN\textsubscript{sim}} = 0.1), where the \RE is now not restricted to being a noun. We replace nouns, verbs, and adjectives in the text similar to the \RE with those extracted from the associated reference review set.

Lastly, we implement W2V-STEM using Gensim~\cite{rehurek_lrec}. We train uni-gram Word2Vec models for single word \REs, and four-gram models for phrases. Models are trained on the training sets. We use cosine similarity to determine the most similar word/phrase in the input text to \RE, which is the replaced \OE. For all other words/phrases, we calculate $w_{i}' = w_{i} - w_{OE} + w_{RE}$, where $w_{i}$ is the original word/phrase's embedding vector, $w_{OE}$ is the \OE's, $w_{RE}$ is the \RE's, and $w_{i}'$ is the resulting embedding vector. The replacement word/phrase is $w_{i}'$'s nearest neighbour. We use similarity thresholds to adjust replacement rates (RR) and produce text under various replacement rate thresholds (RRT).

\section{Evaluation}
\label{sec:results}
\subsection{Evaluation Setup}
%\subsubsection{Selection of Test Entities and Lines    }
We manually select 10 nouns, 10 verbs, 10 adjectives, and 5 phrases from the top 10\% most frequent words/phrases in each test set as our evaluation \REs. We filter the verbs and adjectives through a list of sentiment words~\cite{hu2004mining} to ensure we do not choose \REs that would obviously significantly alter the text's sentiment.\footnote{A list of the chosen \REs along with more detailed explanation of how they were selected is in Appendix D}
%To setup evaluation of our overall pipeline, we select ten keywords or phrases present within each dataset to act as our REs. To do so, we iterate through our test set for each dataset and with the help of the Stanford Parser, we extract a list of nouns and noun phrases. We sort this list by frequency, and limit our selections to the top 10\% most frequent. From these, we manually select ten that are significant and carry strong meaning as the REs for evaluation purposes.
%The reason we select nouns and noun phrases are because they are typically the keywords or key phrases within text. We choose from the most frequent as they are more common and likely hold more significant meaning compared to less frequent nouns such as people's names, typos, and so forth. Manual selection was required as some of the most frequent nouns were words that hold little semantic meaning such as ``it'' and ``they''.

For each evaluation \RE, we choose one-hundred lines from the corresponding test set that does not already contain \RE. We choose lines with at least five words, as many with less carry little semantic meaning (e.g. \textit{`Great!', `It is okay'}). For Amazon and Yelp, we choose 50 positive and 50 negative lines per \RE.\footnote{We don't test on neutral reviews as evaluation of accuracy in sentiment is less well defined (i.e. most ``neutral'' reviews actually carry more positive or negative sentiment)} We repeat this process three times, resulting in three sets of 1000 lines per dataset per POS (excluding phrases), and three sets of 500 lines per dataset for phrases. Our final results are averaged metrics over these three sets.

%Note that we do not choose to test on any neutral (or three-star) reviews due to the noise they carry, as previously mentioned. Further, since our training split was imbalanced and included less neutral reviews, our TIM models learn more from positive and negative reviews and are hence less inclined to produce results that replicate neutral reviews. As such, we cannot accurately test the ability of our pipeline to generate neutral reviews. 

%\subsubsection{Varying Masking Rates}
%To test the ability of our pipeline to perform well on a variety of masking rates, we
For SMERTI-Transformer, SMERTI-RNN, and W2V-STEM, we generate four outputs per text for MRT/RRT of 20\%, 40\%, 60\%, and 80\%, which represent upper-bounds on the percentage of the input that can be masked and/or replaced. Note that NWN-STEM and GWN-STEM can only evaluate on limited POS and their maximum replacement rates are limited.\footnote{See Appendix C for explanations} We select \textit{MIN\textsubscript{sim}} values of 0.075 and 0 for nouns and 0.1 and 0 for verbs, as these result in replacement rates approximately equal to the actual MR/RR of the other models' outputs for 20\% and 40\% MRT/RRT, respectively.
%More details on this are discussed earlier in Section~\ref{sec:MRTST}.
%As these represent upper-bounds, actual masking rates are typically slightly lower. These are reported in the evaluation result tables as Actual MR.
\subsection{Key Evaluation Metrics}
%{\bf Perplexity (PPL)}
%The first metric we use to evaluate the results of our pipeline is perplexity. In particular, we use the news-forward language-model (LM) trained on the One Billion Words corpus by Flair~\cite{akbik2018contextual}. This serves as a strong measurement of the correctness of our outputs in terms of semantics, as outputs which make less sense will result in a higher perplexity since the probability of seeing the output text will be lower. The One Billion Words~\cite{chelba2013one} is a huge text corpus that contains text from a variety of sources, and hence serves as a strong benchmark corpus for our pipeline. Another advantage is that this LM is character-level, and hence our reported perplexity values will be at the character-level. Character-level perplexity is a stronger measurement of correctness, and also handles out-of-vocabulary words including misspellings and typos. We also calculate PPL difference (PPL diff) as the percentage increase or decrease of our output PPL as compared to the input, and include these percentages within our evaluation results.
{\bf Fluency (SLOR)} We use syntactic log-odds ratio (SLOR) \cite{kann-etal-2018-sentence} for sentence level fluency and modify from their word-level formula to character-level ($SLOR_{c}$). We use Flair perplexity values from a language model trained on the One Billion Words corpus \cite{chelba2013one}: %with a bit of mathematical manipulation.

\begin{footnotesize}
\begin{align}
  SLOR_{c}(S)&=\frac{1}{|S|}(ln(p_M(S))-\frac{ln(\prod_{w\in S}p_M(w))}{\sum_{w\in S} |w|}\\
  &\!\!\!\!=-ln(PPL_s)+\frac{\sum_{w\in S} |w| ln(PPL_W)}{\sum_{w\in S} |w|}
  \label{eqn:SLOR}
\end{align}
\end{footnotesize}
where $|S|$ and $|w|$ are the character lengths of the input text $S$ and the word $w$, respectively, $p_M(S)$ and $p_M(w)$ are the probabilities of $S$ and $w$ under the language model $M$, respectively, and $PPL_S$ and $PPL_w$ are the character-level perplexities of $S$ and $w$, respectively. SLOR (from hereon we refer to character-level SLOR as simply SLOR) measures aspects of text fluency such as grammaticality. Higher values represent higher fluency. 

We rescale resulting SLOR values to the interval [0,1] by first fitting and normalizing a Gaussian distribution. We then truncate normalized data points outside [-3,3], which shifts approximately 0.69\% of total data. Finally, we divide each data point by six and add 0.5 to each result.

{\bf Sentiment Preservation Accuracy (SPA)} is defined as the percentage of outputs that carry the same sentiment as the input. We use VADER~\cite{hutto2014vader} to evaluate sentiment as positive, negative, or neutral. It handles typos, emojis, and other aspects of online text. % While our ground truths for Yelp and Amazon already have labelled sentiments (500 positive reviews and 500 negative reviews), in order to accurately assess the sentiment preservation of our pipeline, we run VADER on our ground truth statements as well, and compare the ground truth VADER labels to the VADER labels of associated outputs.

{\bf Content Similarity Score (CSS)} ranges from 0 to 1 and indicates the semantic similarity between generated text and the \RE. A value closer to 1 indicates stronger semantic exchange, as the output is closer in semantic content to the \RE. We also use the USE for this due to its design and strong performance as previously mentioned. %Note that a model which generates text with lower average CSS than the input fails at STE since it alters the text to be more different from the RE.

\subsection{Semantic Text Exchange Score (STES)}
We come up with a single score to evaluate overall performance of a model on STE that combines the key evaluation metrics. It uses the harmonic mean, similar to the F\textsubscript{1} score (or F-score)~\cite{chinchor1992muc, Rijsbergen:1979:IR:539927}, and we call it the {\em Semantic Text Exchange Score (STES)}:
\begin{align}
  STES=\frac{3*A*B*C}{A*B+A*C+B*C}
  \label{eqn:STES}
\end{align}
where $A$ is SPA, $B$ is SLOR, and $C$ is CSS. STES ranges between 0 and 1, with scores closer to 1 representing higher overall performance. Like the F\textsubscript{1} score, STES penalizes models which perform very poorly in one or more metrics, and favors balanced models achieving strong results in all three. %Further, STES being a single score makes it easy to understand and use to compare models. 
%We also introduce a weighted variation of this score called WSTES in Section~\ref{sec:Analysis}.

\subsection{Automatic Evaluation Results}
%probably remove
%We evaluate our two pipeline variations: SMERTI-Transformer and SMERTI-RNN. To the best of our knowledge, while there are competitor models designed to perform specific components of our pipeline such as text infilling, named entity recognition, and semantic similarity and embeddings, none exist for the overall task of semantic text exchange, at least in a focused and rigorous manner. Hence, we do not benchmark directly against any competitors.
%\subsubsection{Overall Averages}
%%%%%%%%%%%%%%%%%%5
%%%INSERT TABLES HERE
%%%%%%%%%%%%%%%%%
%The bolded numbers in the above tables show which of the SMERTI-Transformer and SMERTI-RNN pipelines perform better on that particular metric for the given MR threshold. As stated earlier, either high or low BLEU scores can be desired, so we do not compare their values directly and none are bolded.

\begin{table}
\begin{tabular}{ccc}
\multicolumn{1}{c}{\includegraphics[width=0.45\textwidth]{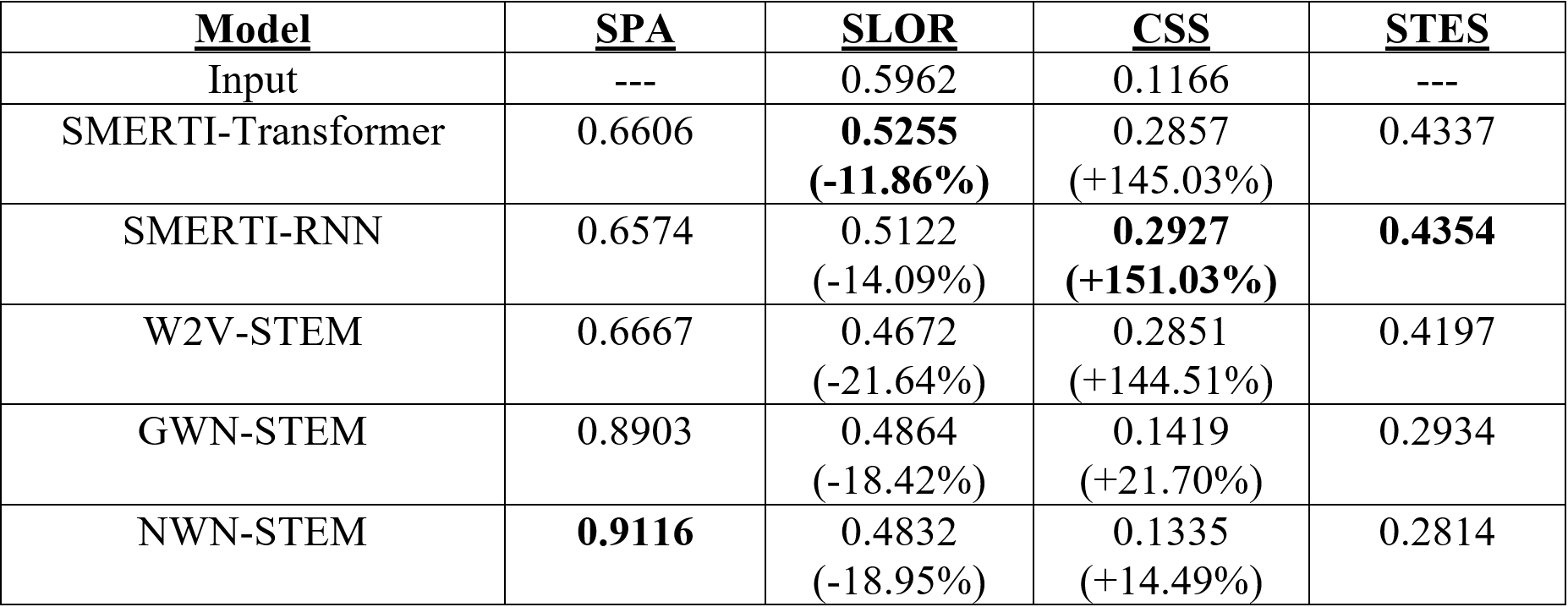}}\\
\end{tabular}
  \caption{\label{tab:results} Overall average results by model (with \% changes from the input)}
\end{table}

%We decide to plot SLOR, BLEU, and SPA, as those are the three most important metrics in evaluating our model.
%See Appendix E for more tables and graphs illustrating key evaluation results per individual dataset.
%Perplexity is on a different scale than the other metrics, and is inherently included in our SLOR score since SLOR is a mathematical manipulation of perplexity and also encompasses other aspects of text fluency.
Table~\ref{tab:results} shows overall average results by model.\footnote{See Appendix E for tables and graphs of detailed results broken down by POS, dataset, and MRT/RRT} Table~\ref{tab:tab2} shows outputs for a Yelp example.\footnote{\label{note1}See Appendix F for many more example outputs from each model for various POS and datasets}
%. See Appendix F for more examples (including for the Amazon reviews and news headlines datasets).
\begin{table*}[t!]
  \centering
  \begin{small}
\begin{tabular}{p{1.2cm}p{12cm}}
    \multicolumn{2}{l}{\textbf{Input text}: great food , large portions ! my family and i really enjoyed our saturday morning breakfast .}\\
\multicolumn{2}{l}{\textbf{Replacement entity}: pizza}\\ \hline
\textbf{MRT/RRT} & \textbf{Generated Output}\\ 
\multicolumn{2}{l}{\underline{SMERTI-Transformer}}\\
    20\%                                                      & great pizza , large slices ! my family and i really enjoyed our saturday morning lunch .\\
  40\%,60\%                                                      & great pizza , large slices ! service was terrific and i really enjoyed our saturday morning lunch .\\
%  60\%                                                      & i love this place! very nice people running the service and the food is always good. food is delicious!\\
  80\%                                                      & great pizza , chewy crust ! nice ambiance and i really enjoyed it . \\
\multicolumn{2}{l}{\underline{SMERTI-RNN}}\\
  20\%                                                      & great pizza , large delivery ! my family and i really enjoyed our saturday morning place .\\
  40\%,60\%                                                      & great pizza , large delivery ! good beer and i really enjoyed our saturday morning place .\\
%  60\%                                                      & i love this place! very nice people running the service and food is always good. thanks!\\
80\%                                                      & great pizza , amazing pizza ! reasonable and i really enjoyed everyone . \\ 
\multicolumn{2}{l}{\underline{W2V-STEM}}\\
    20\%                                                     & great pizza , large portions ! my family and i really enjoyed our saturday morning breakfast .\\
    40\%                                                     & great pizza , large slices ! my family dough i crust enjoyed our saturday morning breakfast .\\
  60\%                                                     & awesome pizza , large slices ! my mom dough i crust enjoyed our saturday morning bagel .\\
  80\%                                              & awesome pizza , slices slices ! my mom dough we crust liked our sunday morning bagel .\\
\multicolumn{2}{l}{\underline{GWN / NWN-STEM}}\\
    20\%                                                      & great food , large stuff ! my family and i really enjoyed our saturday i breakfast .\\
  40\%                                                   & great food , large stuff ! my i and i really enjoyed our saturday i breakfast .\\
\end{tabular}
  \end{small}
\caption{\label{tab:tab2}Generated output text by model for various masking rates on a Yelp evaluation example
  }
\end{table*}

As observed from Table~\ref{tab:tab2} (see also Appendix F), SMERTI is able to generate high quality output text similar to the \RE while flowing better than other models' outputs. It can replace entire phrases and sentences due to its variable length infilling. Note that for nouns, the outputs from GWN-STEM and NWN-STEM are equivalent.\footnote{See Appendix C for explanations}

\subsection{Human Evaluation Setup}
We conduct a human evaluation with eight participants, 6 males and 2 females, that are affiliated project researchers aged 20-39 at the University of Waterloo.\footnote{The authors are not part of the human evaluation} We randomly choose one evaluation line for a randomly selected word or phrase for each POS per dataset. The input text and each model's output (for 40\% MRT/RRT - chosen as a good middle ground) for each line is presented to participants, resulting in a total of 54 pieces of text, and rated on the following criteria from 1-5:
\begin{para_packed_item}
\item \textit{RE Match}: \textit{``How related is the entire text to the concept of [X]", where [X] is a word or phrase (1 - not at all related, 3 - somewhat related, 5 - very related)}. Note here that [X] is a given \RE.
\item \textit{Fluency}: \textit{``Does the text make sense and flow well?" (1 - not at all, 3 - somewhat, 5 - very)}
\item \textit{Sentiment}: \textit{``How do you think the author of the text was feeling?" (1 - very negative, 3 - neutral, 5 - very positive)}
\end{para_packed_item}
Each participant evaluates every piece of text. They are presented with a single piece of text at a time, with the order of models, POS, and datasets completely randomized.

\subsection{Human Evaluation Results}

Average human evaluation scores are displayed in Table~\ref{tab:human_eval_results}. \textit{Sentiment Preservation} (between 0 and 1) is calculated by comparing the average \textit{Sentiment} rating for each model's output text to the \textit{Sentiment} rating of the input text, and if both are less than 2.5 (negative), between 2.5 and 3.5 inclusive (neutral), or greater than 3.5 (positive), this is counted as a valid case of \textit{Sentiment Preservation}. We repeat this for every evaluation line to calculate the final values per model. Harmonic means of all three metrics (using rescaled 0-1 values of \textit{RE Match} and \textit{Fluency}) are also displayed. 
\begin{table}
\begin{tabular}{ccc}
\multicolumn{1}{c}{\includegraphics[width=0.45\textwidth]{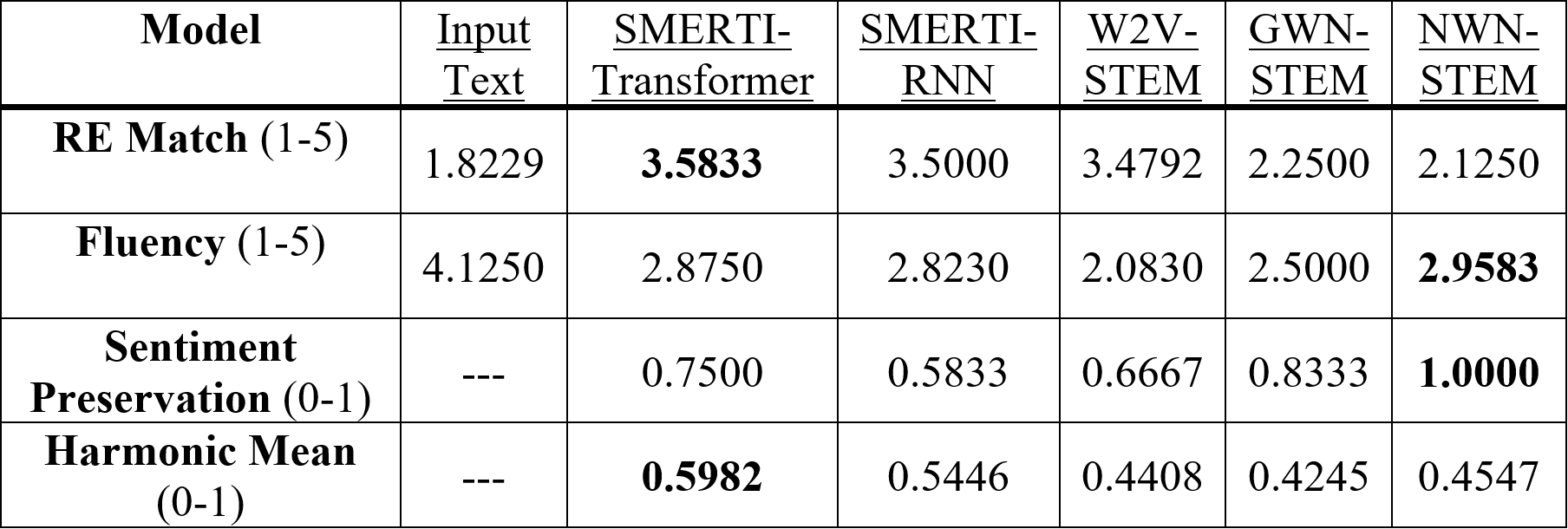}}\\
\end{tabular}
  \caption{\label{tab:human_eval_results} Average human evaluation scores by model}
\end{table}

\section{Analysis}
\subsection{Performance by Model%\footnote{See Appendix I for more detailed analysis on the models}
}
As seen in Table~\ref{tab:results}, both SMERTI variations achieve higher STES and outperform the other models overall, with the WordNet models performing the worst. SMERTI excels especially on fluency and content similarity. The transformer variation achieves slightly higher SLOR, while the RNN variation achieves slightly higher CSS. %Average CSS for SMERTI's output is around 2.5 times that of the original input text.

The WordNet models perform strongest in sentiment preservation (SPA), likely because they modify little of the text and only verbs and nouns. They achieve by far the lowest CSS, %(only around 15-20\% increase compared to input), 
likely in part due to this limited text replacement. They also do not account for context, and many words (e.g. proper nouns) do not exist in WordNet. Overall, the WordNet models are not very effective at STE.

W2V-STEM achieves the lowest SLOR, %(over 20\% drop compared to input text), 
especially for higher RRT, as supported by the example in Table~\ref{tab:tab2} (see also Appendix F). W2V-STEM and WordNet models output grammatically incorrect text that flows poorly. In many cases, words are repeated multiple times. We analyze the average Type Token Ratio (TTR) values of each model's outputs, which is the ratio of unique divided by total words. As shown in Table~\ref{tab:TTR}, the SMERTI variations achieve the highest TTR, while W2V-STEM and NWN-STEM the lowest. 
\begin{table}
\begin{tabular}{ccc}
\multicolumn{1}{c}{\includegraphics[width=0.45\textwidth]{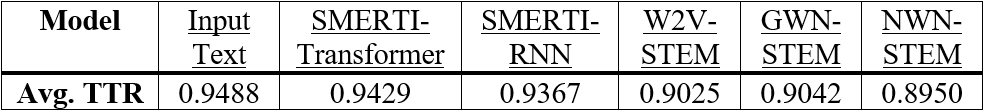}}\\
\end{tabular}
  \caption{\label{tab:TTR} Average TTR values by model}
\end{table}

Note that while W2V-STEM achieves lower CSS than SMERTI, it performs comparably in this aspect. This is likely due to its vector arithmetic operations algorithm, which replaces each word with one more similar to the RE. This is also supported by the lower TTR, as W2V-STEM frequently outputs the same words multiple times.

\subsection{Performance By Model - Human Results}
As seen in Table~\ref{tab:human_eval_results}, the SMERTI variations outperform all baseline models overall, particularly in \textit{RE Match}. SMERTI-Transformer performs the best, with SMERTI-RNN second. The WordNet models achieve high \textit{Sentiment Preservation}, but much lower on \textit{RE Match}. W2V-STEM achieves comparably high \textit{RE Match}, but lowest \textit{Fluency}. 

These results correspond well with our automatic evaluation results in Table~\ref{tab:results}. We look at the Pearson correlation values between \textit{RE Match}, \textit{Fluency}, and \textit{Sentiment Preservation} with CSS, SLOR, and SPA, respectively. These are 0.9952, 0.9327, and 0.8768, respectively, demonstrating that our automatic metrics are highly effective and correspond well with human ratings.

\subsection{SMERTI's Performance By POS}

\begin{table}
\begin{tabular}{ccc}
\multicolumn{1}{c}{\includegraphics[width=0.45\textwidth]{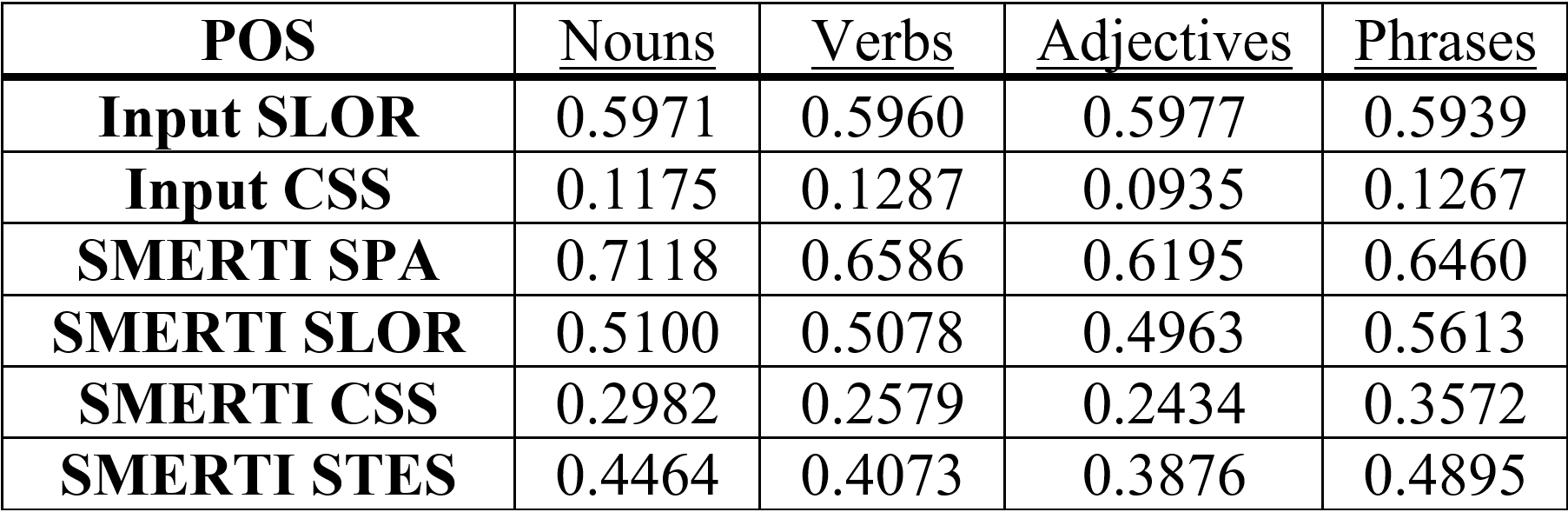}}\\
\end{tabular}
  \caption{\label{tab:POS_results} Input text's avg. SLOR, CSS, and SMERTI's avg. SPA, SLOR, CSS, and STES by POS}
\end{table}

As seen from Table~\ref{tab:POS_results}\footnote{Note that the SMERTI values in Tables~\ref{tab:POS_results} to~\ref{tab:avg_results} refer to the average between SMERTI-Transformer and SMERTI-RNN} , SMERTI's SPA values are highest for nouns, likely because they typically carry little sentiment, and lowest for adjectives, likely because they typically carry the most. %, even though we filtered our evaluation REs through a list of sentiment words.\footnote{Consider words such as ``small", ``new", and ``busy", which are not directly associated with sentiment. However, they can lean towards positive or negative (i.e. ``small" may be commonly associated with ``small portions" to describe negative experiences at restaurants)} 

%\footnote{See Appendix K for more on SPA vs. POS} %even though we filtered our evaluation REs through a list of sentiment words. To see why, consider words such as ``small", ``new", and ``busy". While these words themselves are not directly associated with positive or negative sentiment, they can sway one way or the other, particular in their associated datasets.\footnote{i.e. ``small" may be commonly associated with ``small portions" to describe negative experiences at restaurants}

SLOR is lowest for adjectives and highest for phrases and nouns. Adjectives typically carry less semantic meaning and SMERTI likely has more trouble figuring out how best to infill the text. In contrast, nouns typically carry more, and phrases the most (since they consist of multiple words).

SMERTI's CSS is highest for phrases then nouns, likely due to phrases and nouns carrying more semantic meaning, making it easier to generate semantically similar text. Both SMERTI's and the input text's CSS are lowest for adjectives, likely because they carry little semantic meaning.

Overall, SMERTI appears to be more effective on nouns and phrases than verbs and adjectives.

\subsection{SMERTI's Performance By Dataset}

\begin{table}
\begin{tabular}{ccc}
\multicolumn{1}{c}{\includegraphics[width=0.45\textwidth]{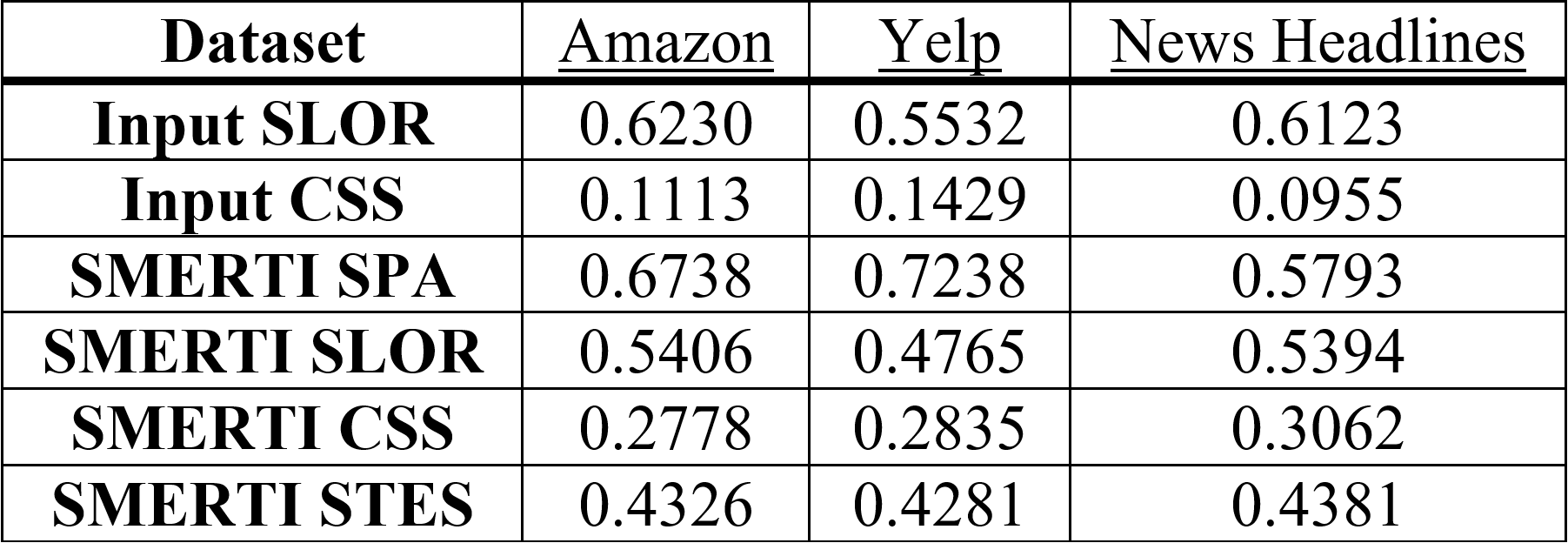}}\\
\end{tabular}
  \caption{\label{tab:dataset_results} Input text's avg. SLOR, CSS, and SMERTI's avg. SPA, SLOR, CSS, and STES by dataset}
\end{table}

As seen in Table~\ref{tab:dataset_results}, SMERTI's SPA is lowest for news headlines. Amazon and Yelp reviews naturally carry stronger sentiment, likely making it easier to generate text with similar sentiment.

Both SMERTI's and the input text's SLOR appear to be lower for Yelp reviews. This may be due to many reasons, such as more typos and emojis within the original reviews, and so forth.

SMERTI's CSS values are slightly higher for news headlines. This may be due to them typically being shorter and carrying more semantic meaning as they are designed to be attention grabbers. 

Overall, it seems that using datasets which inherently carry more sentiment will lead to better sentiment preservation. Further, the quality of the dataset's original text, unsurprisingly, influences the ability of SMERTI to generate fluent text. %Lastly, stronger content similarity appears easier to achieve on text that is initially shorter and carries more semantic meaning.

\subsection{SMERTI's Performance By MRT/RRT}
\label{sec:analysis}

\begin{table}
\begin{tabular}{ccc}
\multicolumn{1}{c}{\includegraphics[width=0.45\textwidth]{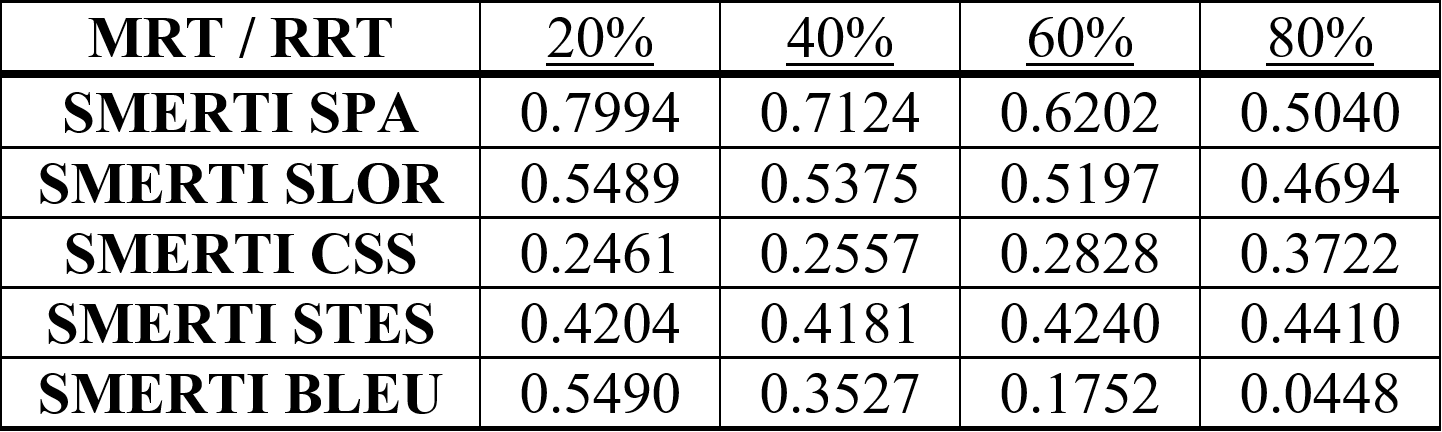}}\\
\end{tabular}
  \caption{\label{tab:avg_results} SMERTI's avg. SPA, SLOR, CSS, STES, and BLEU by MRT/RRT}
\end{table}

From Table~\ref{tab:avg_results}, it can be seen that as MRT/RRT increases, SMERTI's SPA and SLOR decrease while CSS increases. These relationships are very strong as supported by the Pearson correlation values of -0.9972, -0.9183, and 0.9078, respectively. When SMERTI can alter more text, it has the opportunity to replace more related to sentiment while producing more of semantic similarity to the \RE.

Further, SMERTI generates more of the text itself, becoming less similar to the human-written input, resulting in lower fluency. To further demonstrate this, we look at average SMERTI BLEU ~\cite{papineni2002} scores against MRT/RRT, shown in Table~\ref{tab:avg_results}. BLEU generally indicates how close two pieces of text are in content and structure, with higher values indicating greater similarity. We report our final BLEU scores as the average scores of 1 to 4-grams. As expected, BLEU decreases as MRT/RRT increases, and this relationship is very strong as supported by the Pearson correlation value of -0.9960.
%Note that BLEU has a brevity penalty, and some of the output lines with 80\% MRT are noticeably shorter than the original text, leading to much lower BLEU scores.

%SMERTI really shines on higher thresholds, exceeding W2V-STEM on nearly every metric for 60\% and 80\% MRT/RRT (???). This is evident from the example in Table~\ref{tab:tab2} (see also Appendix G). SMERTI has higher SLOR across the board, and this difference increases with MRT/RRT.

It is clear that MRT/RRT represents a trade-off between CSS against SPA and SLOR. It is thus an adjustable parameter that can be used to control the generated text, and balance semantic exchange against fluency and sentiment preservation.

\section{Conclusion and Future Work}
We introduced the task of semantic text exchange (STE), demonstrated that our pipeline SMERTI performs well on STE, and proposed an STES metric for evaluating overall STE performance. SMERTI outperformed other models and was the most balanced overall. %across the various parts-of-speech, datasets, and masking (replacement) rate thresholds (MRT/RRT). 
We also showed a trade-off between semantic exchange against fluency and sentiment preservation, which can be controlled by the masking (replacement) rate threshold.

Potential directions for future work include adding specific methods to control sentiment, and fine-tuning SMERTI for preservation of persona or personality. Experimenting with other text infilling models (e.g. fine-tuning BERT~\cite{devlin-etal-2019-bert}) is also an area of exploration. Lastly, our human evaluation is limited in size and a larger and more diverse participant pool is needed.

We conclude by addressing potential ethical misuses of STE, including assisting in the generation of spam and fake-reviews/news. These risks come with any intelligent chatbot work, but we feel that the benefits, including usage in the detection of misuse such as fake-news, greatly outweigh the risks and help progress NLP and AI research.

\section*{Acknowledgments}
We thank our anonymous reviewers, study participants, and Huawei Technologies Co., Ltd. for financial support.
\bibliography{emnlp-ijcnlp-2019}
\bibliographystyle{acl_natbib}

%\appendix
%\section{Supplemental Material}
%\label{sec:supplemental}

\end{document}

\maketitle
\footnotetext[1]{Authors contributed equally}

\renewcommand*{\thefootnote}{\arabic{footnote}}

\appendix
\section{Masked Output Examples}
\label{sec:appendix_A}

Table\footnote{Tables and figures mentioned in this appendices document refer to the tables and figures here}~\ref{tab:masked_outputs} includes various example masked outputs by the ERM and SMM modules. We illustrate a variety of word-to-word, phrase-to-word, word-to-phrase, and phrase-to-phrase entity replacements and similarity masking.

\begin{table*}[t!]
%  \leftalign
  \begin{footnotesize}
\begin{tabular}{|p{0.5cm}|p{14cm}|}
  \hline
  $S$ & {\em family enjoyed the food quite a bit especially the sweet and sour chicken .} \\
  \RE & {\em bitter} \\
  \hline
  $S''_1$ & {\em family enjoyed the food quite a bit especially the sweet and bitter chicken .} \\
  $S''_2$ & {\em family [mask] the [mask] a bit especially the sweet and bitter [mask] .} \\
  $S''_3$ & {\em [mask] the [mask] bit [mask] the [mask] and bitter [mask] .} \\
  $S''_4$ & {\em [mask] bit [mask] bitter [mask] .}\\
  \Xhline{3\arrayrulewidth}
  
  $S$ & {\em terrible customer service . couldn't make a wire transfer because they are out of paper .} \\
  \RE & {\em amazing} \\
  \hline
  $S''_1$ & {\em amazing customer [mask] . couldn't make a wire transfer [mask] they are out of paper .} \\
  $S''_2$ & {\em amazing customer [mask] . couldn't make a wire transfer [mask] they [mask] out of paper .} \\
  $S''_3$ & {\em amazing [mask] . [mask] make a wire transfer [mask] out [mask] paper .} \\
  $S''_4$ & {\em amazing [mask] . [mask] a [mask] out [mask] .} \\
  \Xhline{3\arrayrulewidth}
  
  $S$ & {\em Heather enjoyed her movie date with Jim last night.} \\
  \RE & {\em yesterday} \\
  \hline
  $S''_1$ & {\em heather [mask] her movie [mask] with jim yesterday .} \\
  $S''_2$ & {\em heather [mask] her movie [mask] with jim yesterday . } \\
  $S''_3$ & {\em heather [mask] her movie [mask] with jim yesterday .} \\
  $S''_4$ & {\em [mask] yesterday .} \\
  \Xhline{3\arrayrulewidth}
  
  $S$ & {\em Heather enjoyed her movie date with Jim last night.} \\
  \RE & {\em Elizabeth} \\
  \hline
  $S''_1$ & {\em elizabeth enjoyed [mask] movie date with [mask] last night .} \\
  $S''_2$ & {\em elizabeth enjoyed [mask] movie date with [mask] .} \\
  $S''_3$ & {\em elizabeth enjoyed [mask] movie date with [mask] .} \\
  $S''_4$ & {\em elizabeth enjoyed [mask] with [mask] .} \\
  \Xhline{3\arrayrulewidth}
  
  $S$ & {\em The car crashed into the building and exploded, killing hundreds.} \\
  \RE & {\em caught on fire} \\
  \hline
  $S''_1$ & {\em the car [mask] caught on fire and [mask] , killing hundreds .} \\
  $S''_2$ & {\em [mask] caught on fire and [mask] , killing hundreds .} \\
  $S''_3$ & {\em [mask] caught on fire and [mask] , killing hundreds .} \\
  $S''_4$ & {\em [mask] caught on fire [mask] .} \\
  \Xhline{3\arrayrulewidth}
  
  $S$ & {\em My son took his math test yesterday and failed. He cried all day and I hate him now.} \\
  \RE & {\em medical examination} \\
  \hline
  $S''_1$ & {\em [mask] took medical examination yesterday and [mask] . he cried all day and i hate [mask] now .} \\
  $S''_2$ & {\em [mask] took medical examination yesterday and [mask] . he cried all day and i hate [mask] now .} \\
  $S''_3$ & {\em [mask] took medical examination yesterday and [mask] . he cried all day and i hate [mask] now .} \\
  $S''_4$ & {\em [mask] medical examination [mask] and [mask] . [mask] .} \\
  \Xhline{3\arrayrulewidth}
  
  $S$ & {\em I took my dog for a walk in the park. He really enjoyed it!} \\
  \RE & {\em the river} \\
  \hline
  $S''_1$ & {\em i took my dog for [mask] in the river . he really enjoyed it !} \\
  $S''_2$ & {\em i took [mask] for [mask] in the river . he really enjoyed it !} \\
  $S''_3$ & {\em i [mask] for [mask] in the river . he [mask] !} \\
  $S''_4$ & {\em i [mask] the river . he [mask] !} \\
  \Xhline{3\arrayrulewidth}
  
  $S$ & {\em It is very sunny outside so I am very sweaty.} \\
  \RE & {\em extremely snowy} \\
  \hline
  $S''_1$ & {\em it is extremely snowy outside so i am [mask] .} \\
  $S''_2$ & {\em it is extremely snowy [mask] so i am [mask] .} \\
  $S''_3$ & {\em it [mask] extremely snowy [mask] i [mask] .} \\
  $S''_4$ & {\em [mask] extremely snowy [mask] .} \\
  \Xhline{3\arrayrulewidth}
  
  $S$ & {\em I went to my friend Amy's house last night.} \\
  \RE & {\em my husband Jim's} \\
  \hline
  $S''_1$ & {\em i went to my husband jim's house last night .} \\
  $S''_2$ & {\em i went to my husband jim's house [mask] .} \\
  $S''_3$ & {\em i went to my husband jim's [mask] .} \\
  $S''_4$ & {\em i [mask] my husband jim's [mask] .} \\
  \Xhline{3\arrayrulewidth}
  
  $S$ & {\em I went to my friend Amy's house last night.} \\
  \RE & {\em my husband Jim's boat} \\
  \hline
  $S''_1$ & {\em i went to my husband jim's boat last night .} \\
  $S''_2$ & {\em i went to my husband jim's boat last night .} \\
  $S''_3$ & {\em i went to my husband jim's boat [mask] .} \\
  $S''_4$ & {\em i [mask] my husband jim's boat [mask] .} \\
  \hline
  
\end{tabular}
  \end{footnotesize}
\caption{\label{tab:masked_outputs}Example masked outputs. $S$ is the original input text; $RE$ is the replacement entity;\\ $S''_1$  corresponds to $MRT = 0.2$, base $ST = 0.4$; $S''_2$ corresponds to $MRT = 0.4$, base $ST = 0.3$;\\ $S''_3$ corresponds to $MRT = 0.6$, base $ST = 0.2$; $S''_4$ corresponds to $MRT = 0.8$, base $ST = 0.1$}
\end{table*}

\section{Dataset Statistics}
\label{sec:appendix_B}

See Table~\ref{tab:dataset_statistics} for our training and testing splits by dataset and sentiment.

\begin{table}
\begin{tabular}{ccc}
\multicolumn{1}{c}{\includegraphics[width=0.45\textwidth]{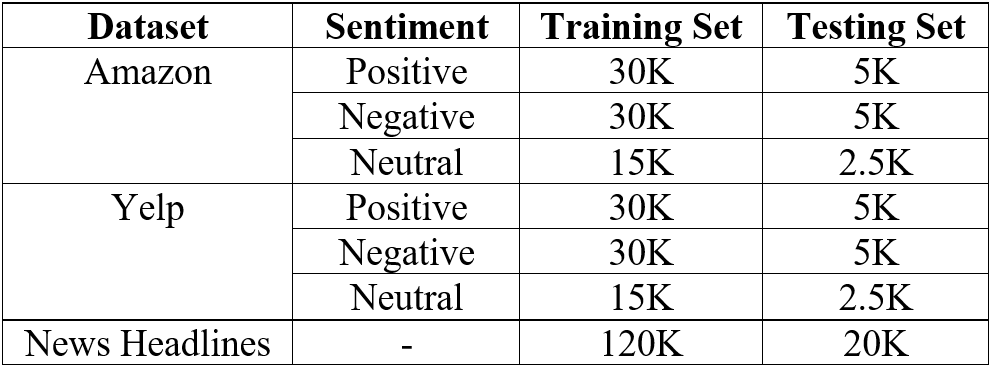}}\\
\end{tabular}
  \caption{\label{tab:dataset_statistics} Training and testing splits by dataset}
\end{table}

\section{Details of Baseline Implementations}

\subsection{NWN-STEM}
This follows Algorithm 2 in \citet{yao2017automated}. For each dataset and each \RE (note that this model is restricted to noun \REs), we go through the dataset's training set (which acts as the ``reference review set") and extract a list of text lines that contain the \RE (where the \RE acts as the ``topic keyword"). For each of these lines, with the help of the Stanford Parser, we extract all single-word nouns, and for each of them (which we call $noun_i$), we check if $MIN_{sim}(noun_i, RE) > 0.1$. If so, we add them to the list of nouns similar to the RE, which we call $sim_{nouns}$. 

For each evaluation line and associated \RE, we extract all singular nouns within the text that are similar to the \RE. For our evaluation purposes, we choose two $MIN_{sim}$ values of 0.075 and 0 to produce two outputs per input. These two $MIN_{sim}$ values result in actual replacement rates similar to the actual masking/replacement rates of other models (SMERTI and W2V-STEM) for MRT/RRT of 20\% and 40\%, respectively. Each similar noun is replaced with the noun in $sim_{nouns}$ that is most similar to it to produce the output text.

When analyzing every word during the above procedure, we take the first default WordNet synset's definition of the word (e.g. ``bill" will default to ``a statute in draft before it becomes law"). Further, if a word does not exist in WordNet, we use WordNet's Morphy to find if a lexically similar word exists that may just be in a different form (e.g. ``photos" does not exist in WordNet, but using Morphy, we can find ``photograph") and use the first default synset of this resulting word.

Note that since the number of nouns per input is limited, the actual replacement rates have an upper limit, and this is why we only generate two outputs per input to compare to outputs of other models for 20\% and 40\% MRT/RRT.

\subsection{GWN-STEM}

This is a modification of NWN-STEM to handle verbs and adjectives (note that WordNet only works for single words). On top of nouns, we also extract similar verbs and adjectives to the \RE. This results in three lists: $sim_{nouns}$, $sim_{verbs}$, and $sim_{adjs}$, where we still use a $MIN_{sim}$ value of 0.1 for determining the three lists. Further, we not only replace nouns in the original text, but also verbs and adjectives. Note that we use the same synset and Morphy procedure as for NWN-STEM. For our evaluation, we choose the same $MIN_{sim}$ values of 0.075 and 0 for noun \REs, but $MIN_{sim}$ values of 0.1 and 0 for verb \REs. These combinations of $MIN_{sim}$ values result in actual replacement rates similar to the actual masking/replacement rates of other models for MRT/RRT of 20\% and 40\%, respectively.

We noticed that GWN-STEM only works for noun and verb \REs, as for most adjectives, WordNet cannot calculate similarity scores. Hence, it was infeasible to evaluate on adjective \REs. Further, most similarity scores only exist between noun-noun pairs and verb-verb pairs, and when we tried to produce $sim_{verbs}$ and $sim_{adjs}$ for noun \REs, almost all resulted in empty lists. Hence, GWN-STEM actually produced the same outputs as NWN-STEM for noun \REs (and was also limited to two replacement rates), which is why the two models have the same outputs and resulting metrics for noun \REs. Even for verb \REs, we were limited to two sets of outputs (mimicking the two replacement rates above) since similarity calculations between verb-noun pairs and verb-adjective pairs were limited, so few were replaced.

\subsection{W2V-STEM}
This uses Word2Vec (W2V) models trained using Gensim. We train six W2V models: one unigram model per dataset, and one four-gram model per dataset, where each is trained using the corresponding dataset's training set. To train the four-gram models, we begin by applying a bi-gram phrasing model on top of the original text, and then the bi-gram phrasing model again on top of this resulting text. We call this a four-gram phrasing model. We then use this to generate text that is grouped into phrases up to four-grams long. We then train W2V models on this four-gram text to generate the four-gram W2V models.

For the unigram models, we use an embedding vector size of 50, a context window of 3, a minimum token count of 0, and the skip-gram model. For the four-gram models, we use an embedding vector size of 10, a context window of 1, a minimum token count of 0, and the CBOW (continuous bag-of-words) model.

For evaluation lines with noun, verb, and adjective \REs, we go through the line of text and with the help of the Stanford parser, extract all words that are the same POS as the \RE (which become the candidate \OEs). Then, using cosine similarity between the W2V embedding vectors (from the unigram W2V models) of the \RE and candidate \OEs, we find the word with the maximum similarity to the \RE which becomes the replaced \OE. Then, we replace other words in the input text similar to the \OE using vector addition and subtraction as described in Section 4.3. We start with a similarity threshold value of 0.1 that increases by 0.05 each time to generate text satisfying varying replacement rate thresholds.

For evaluation lines with phrase \REs, we first generate text files of the evaluation sets that are grouped into phrases up to four-grams long using the four-gram phrasing model. Then, using the four-gram W2V models, we proceed similarly as above to determine the replaced \OE, which can now be either a single word or a phrase. Other words and phrases in the input text are replaced using vector addition and subtraction. We start with a similarity threshold value of 0.3 that increases by 0.01 each time to generate text satisfying varying replacement rate thresholds.

\section{Evaluation Keywords/Phrases}

We select ten keywords (five keyphrases) to act as our \REs per dataset per POS. See Tables~\ref{tab:evaluation_nouns} to~\ref{tab:evaluation_phrases} for the chosen \REs.  To do so, we iterate through our test sets and with the Stanford Parser, we extract a list of nouns and noun phrases, verbs and verb phrases, and adjectives and adjective phrases. We sort these lists by frequency, and limit our selections to the top 10\% most frequent. For the verbs and adjectives and their phrases, we further filter them through a list of sentiment words \cite{hu2004mining} to ensure the \REs we choose do not carry significant sentiment-related meaning, as inserting them into the original text would obviously lead to major changes in sentiment. From these, we manually select ten per dataset per POS (except phrases, where we select five per dataset) that are significant and carry strong meaning. These work well as the \REs for evaluation purposes. Note that for phrases, we choose three noun phrases, one verb phrase, and one adjective phrase per dataset.

We choose from the most frequent words and phrases as they are more common and likely hold more significant meaning compared to less frequent ones (e.g. names and typos). Manual selection was required as some of the most frequent words/phrases hold little semantic meaning (e.g. \textit{it}, \textit{they}, \textit{is}, \textit{was}, and so forth). We only choose half the number of phrases as words as we find that frequent phrases carrying significant semantic meaning with little sentiment are much rarer.

\begin{table}
\begin{tabular}{ccc}
\multicolumn{1}{c}{\includegraphics[width=0.45\textwidth]{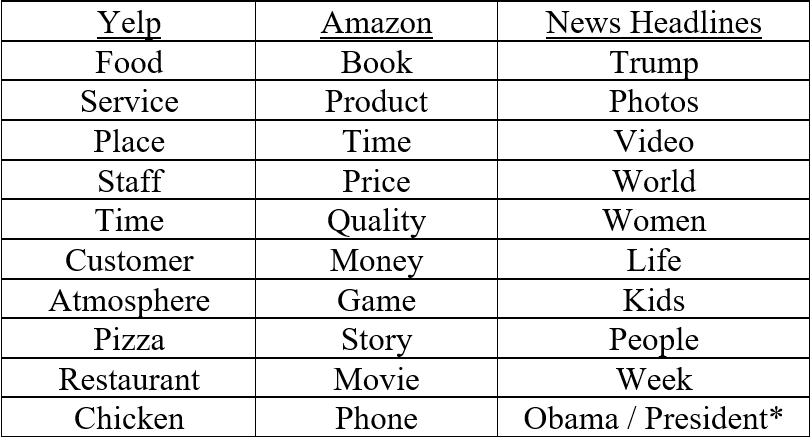}}\\
\end{tabular}
  \caption{\label{tab:evaluation_nouns} Chosen evaluation noun \REs. *\textit{Obama} does not exist in WordNet, so we instead use the word \textit{President} for NWN-STEM and GWN-STEM.}
\end{table}

\begin{table}
\begin{tabular}{ccc}
\multicolumn{1}{c}{\includegraphics[width=0.45\textwidth]{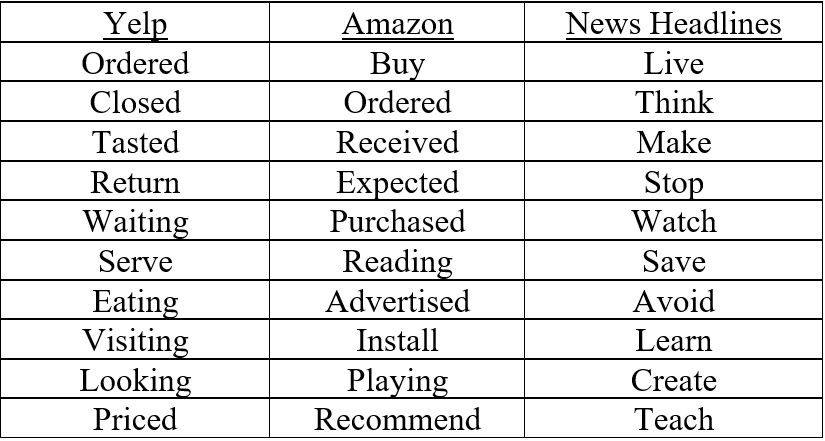}}\\
\end{tabular}
  \caption{\label{tab:evaluation_verbs} Chosen evaluation verb \REs}
\end{table}

\begin{table}
\begin{tabular}{ccc}
\multicolumn{1}{c}{\includegraphics[width=0.45\textwidth]{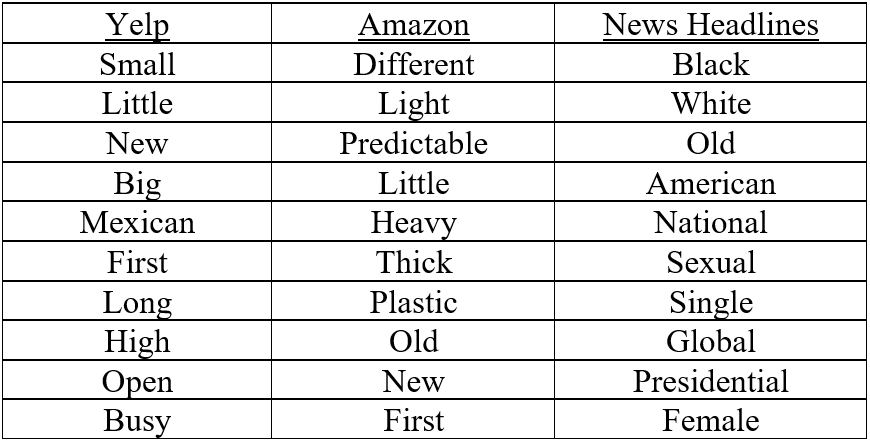}}\\
\end{tabular}
  \caption{\label{tab:evaluation_adjs} Chosen evaluation adjective \REs}
\end{table}

\begin{table}
\begin{tabular}{ccc}
\multicolumn{1}{c}{\includegraphics[width=0.45\textwidth]{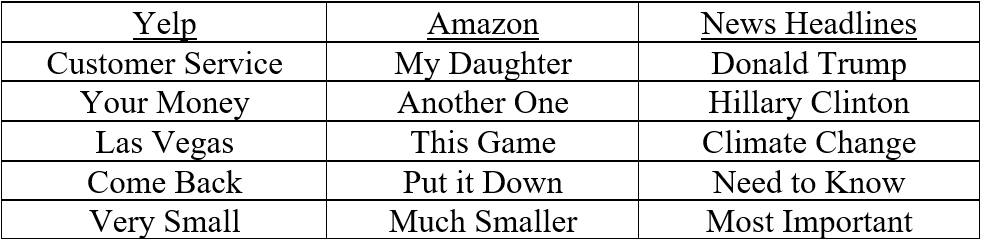}}\\
\end{tabular}
  \caption{\label{tab:evaluation_phrases} Chosen evaluation phrase \REs}
\end{table}

\section{Detailed Evaluation Results - Tables and Graphs}

\begin{figure}
\begin{tabular}{ccc}
\multicolumn{1}{c}{\includegraphics[width=0.45\textwidth]{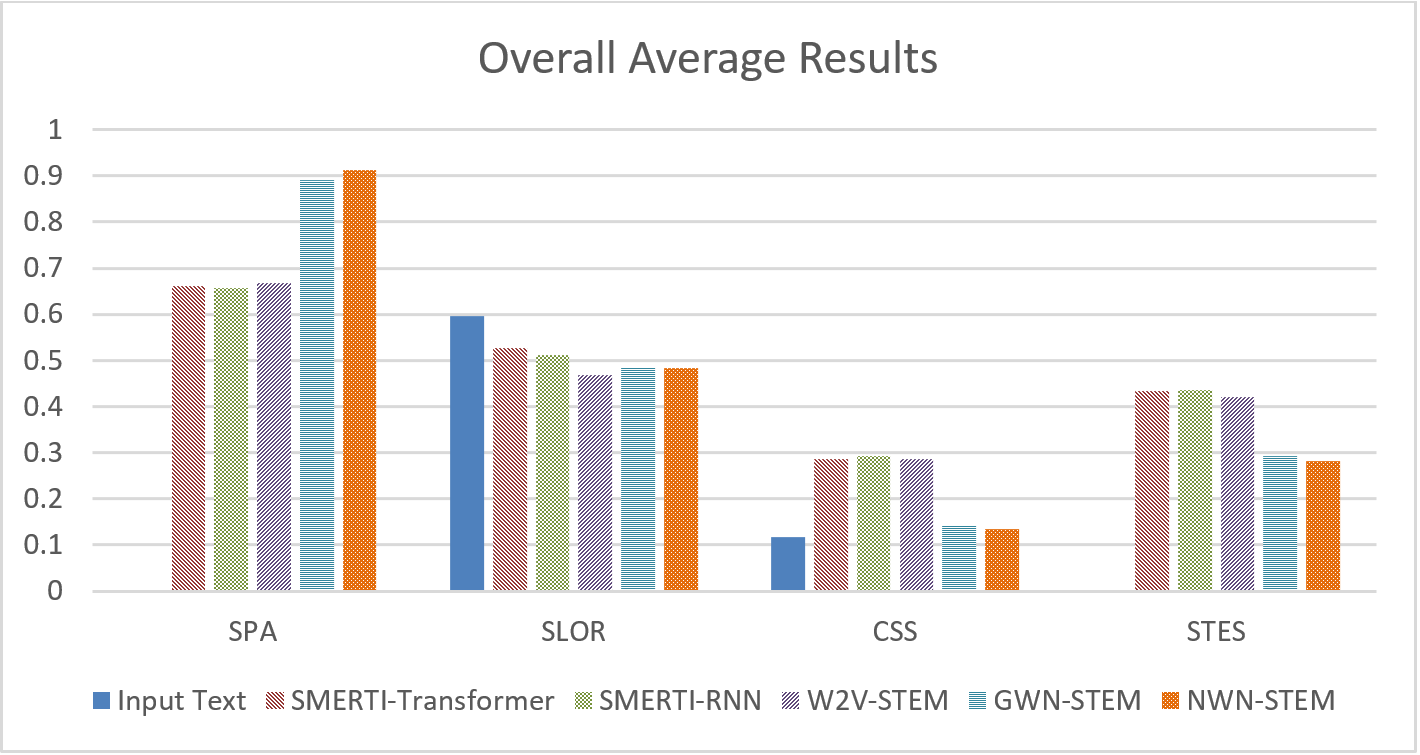}}\\
\end{tabular}
  \caption{\label{fig:overall_results_graph} Graph of overall average results (referring to the data found in Table 2 of the main body)}
\end{figure}

\begin{figure*}
\begin{tabular}{ccc}
\multicolumn{1}{c}{\includegraphics[width=0.95\textwidth]{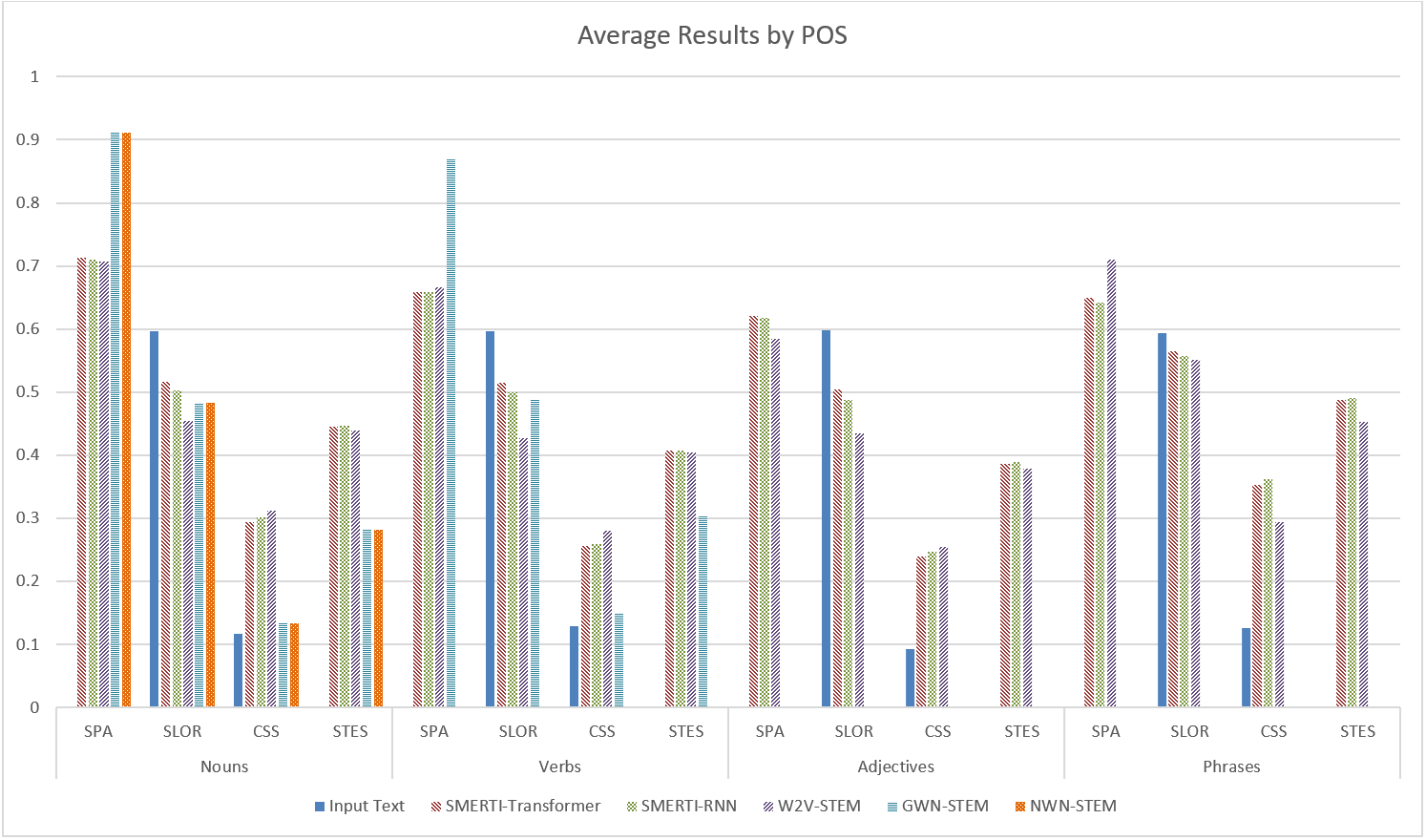}}\\
\end{tabular}
  \caption{\label{fig:POS_results_graph} Graph of average results by POS}
\end{figure*}

\begin{figure*}
\begin{tabular}{ccc}
\multicolumn{1}{c}{\includegraphics[width=0.95\textwidth]{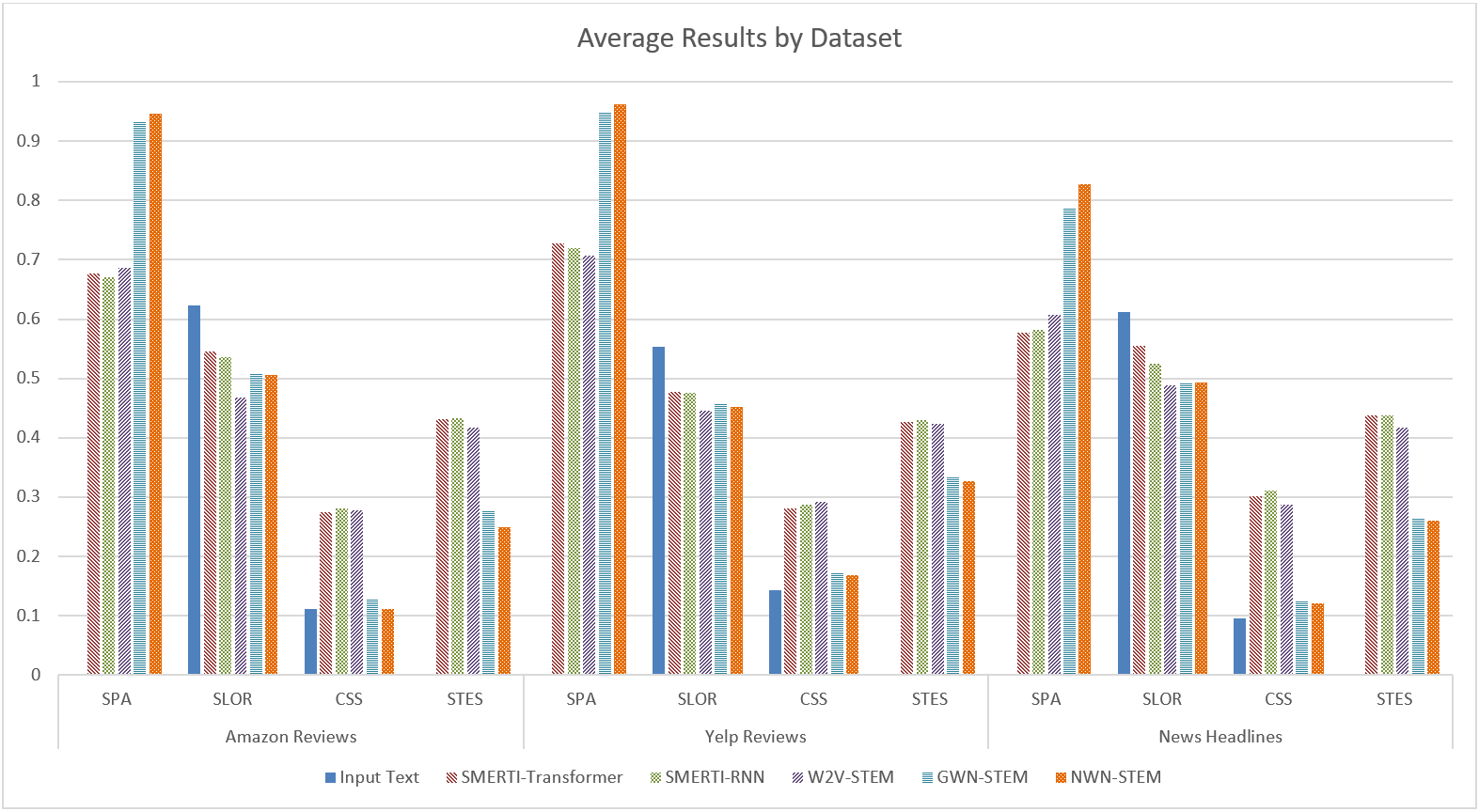}}\\
\end{tabular}
  \caption{\label{fig:dataset_results_graph} Graph of average results by dataset}
\end{figure*}

\begin{figure*}
\begin{tabular}{ccc}
\multicolumn{1}{c}{\includegraphics[width=0.95\textwidth]{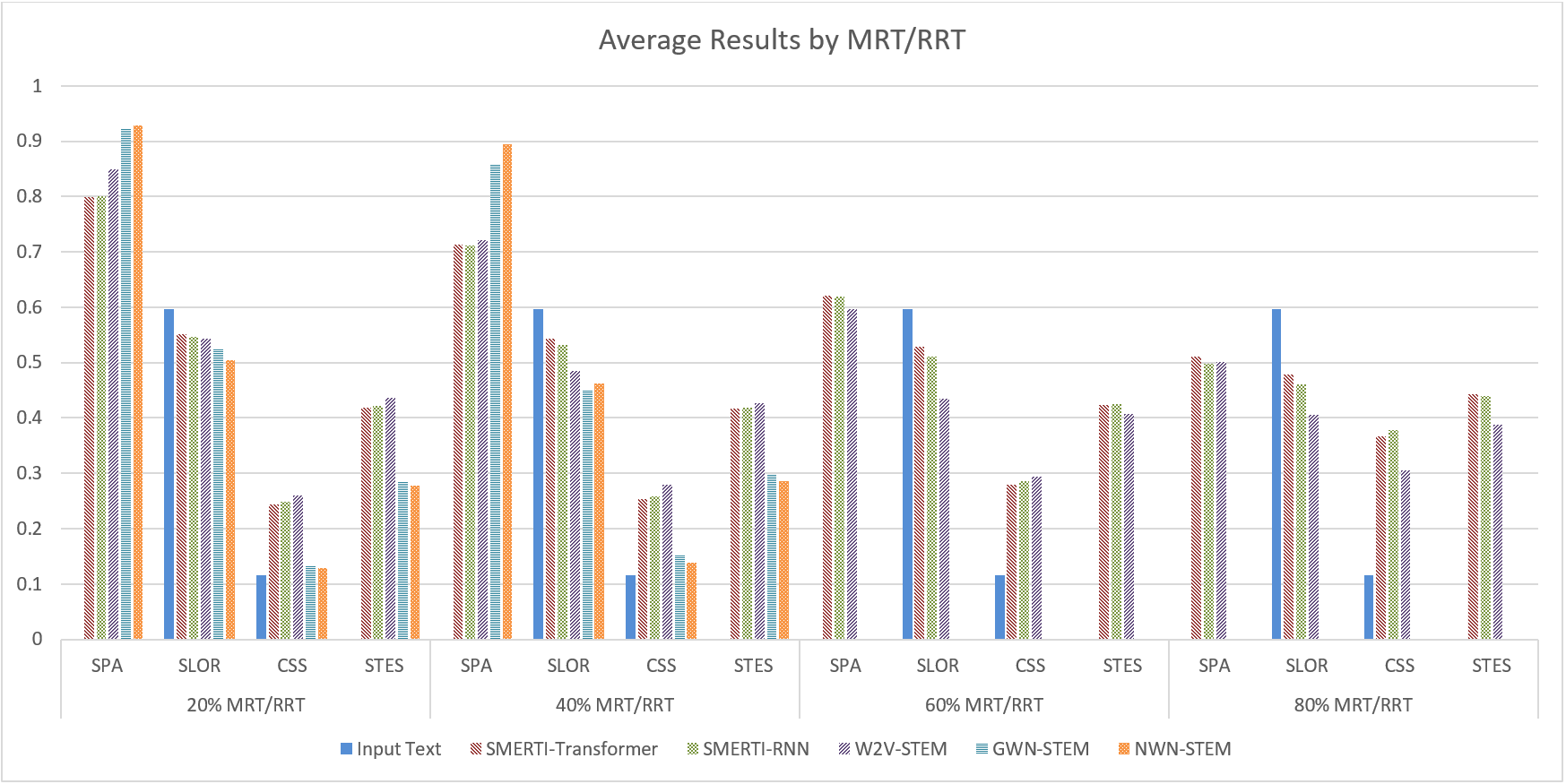}}\\
\end{tabular}
  \caption{\label{fig:MRT_results_graph} Graph of average results by MRT/RRT}
\end{figure*}

See Figure~\ref{fig:overall_results_graph} for a graph of overall average results, Table~\ref{tab:POS_results_table} and Figure~\ref{fig:POS_results_graph} for average results by POS, Table~\ref{tab:dataset_results_table} and Figure~\ref{fig:dataset_results_graph} for average results by dataset, and Table~\ref{tab:MRT_results_table} and Figure~\ref{fig:MRT_results_graph} for average results by MRT/RRT. Note that the bolded values in the tables show which model performs better on that particular metric, on average, for the category. 

\begin{table*}
\begin{tabular}{ccc}
\multicolumn{1}{c}{\includegraphics[width=0.95\textwidth]{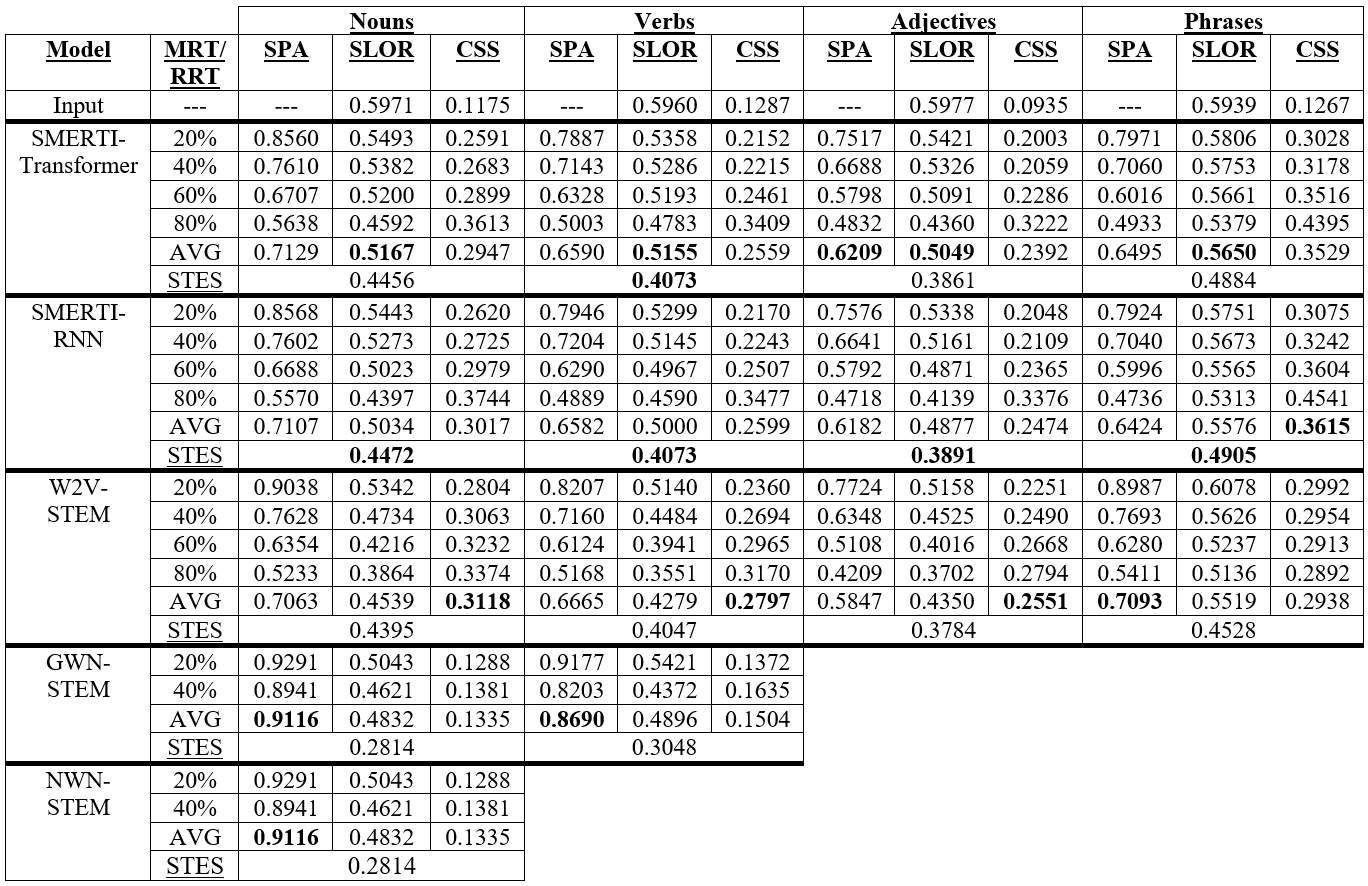}}\\
\end{tabular}
  \caption{\label{tab:POS_results_table} Average results by POS}
\end{table*}

\begin{table*}
\begin{tabular}{ccc}
\multicolumn{1}{c}{\includegraphics[width=0.95\textwidth]{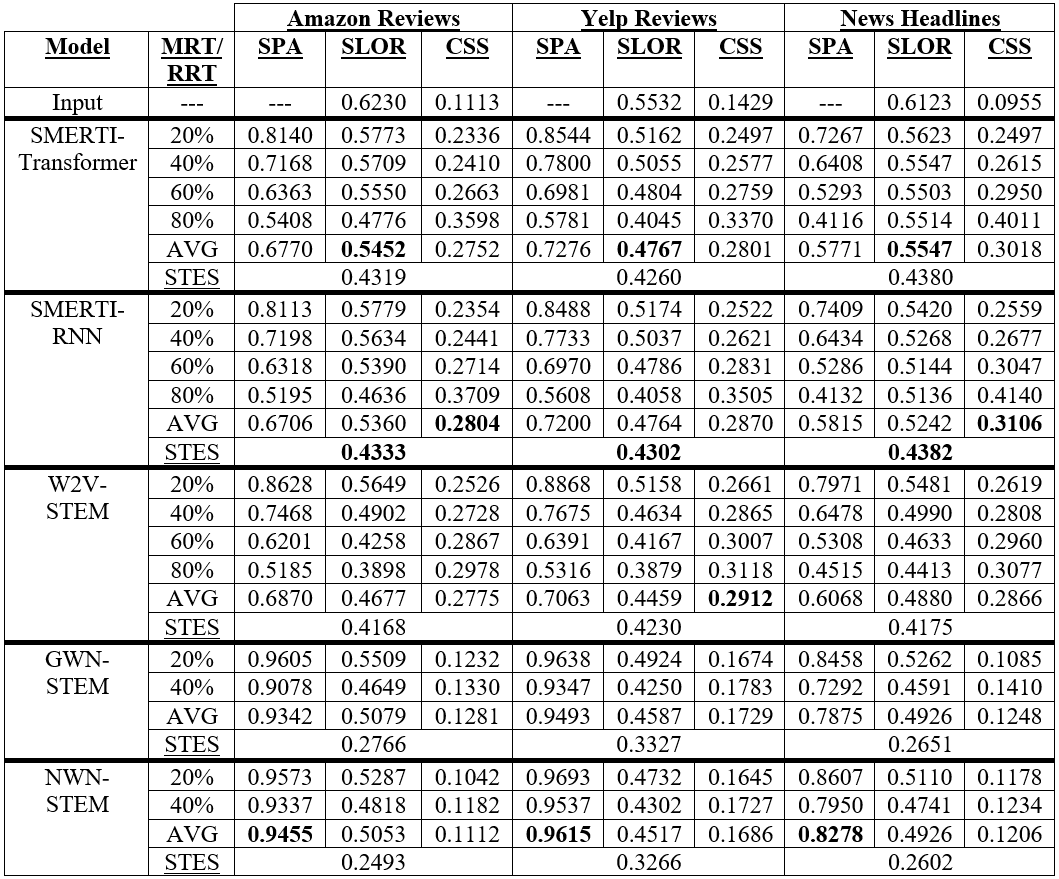}}\\
\end{tabular}
  \caption{\label{tab:dataset_results_table} Average results by dataset}
\end{table*}

\begin{table*}
\begin{tabular}{ccc}
\multicolumn{1}{c}{\includegraphics[width=0.95\textwidth]{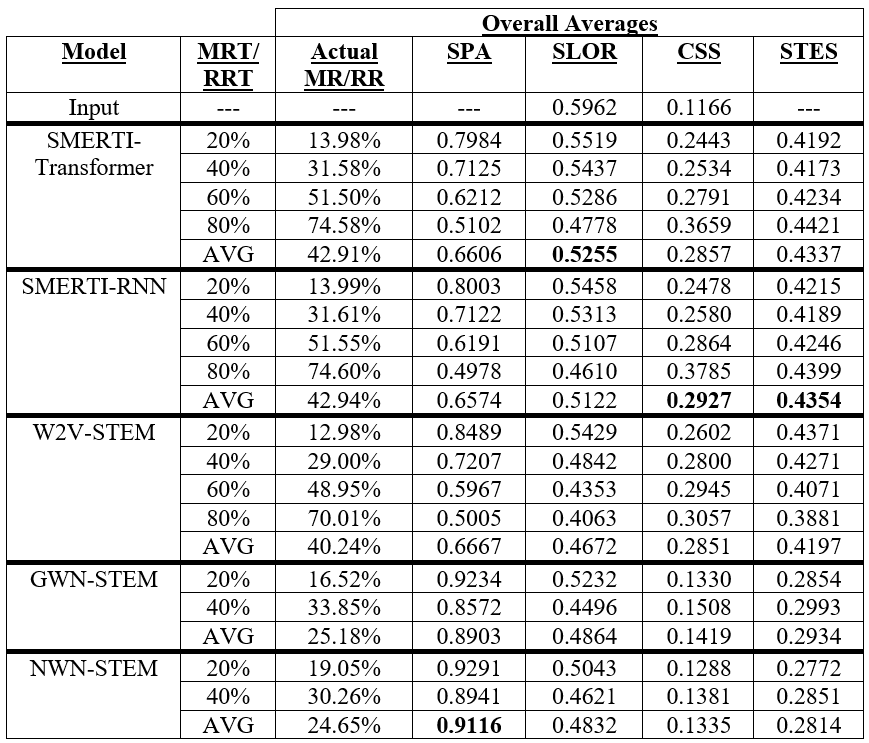}}\\
\end{tabular}
  \caption{\label{tab:MRT_results_table} Average results by MRT/RRT}
\end{table*}

\section{Model Output Examples}

See Tables~\ref{tab:amazon_noun} to~\ref{tab:news_phrase} for example outputs from every model for all datasets and POS.

\begin{table*}
\begin{tabular}{ccc}
\multicolumn{1}{c}{\includegraphics[width=0.95\textwidth]{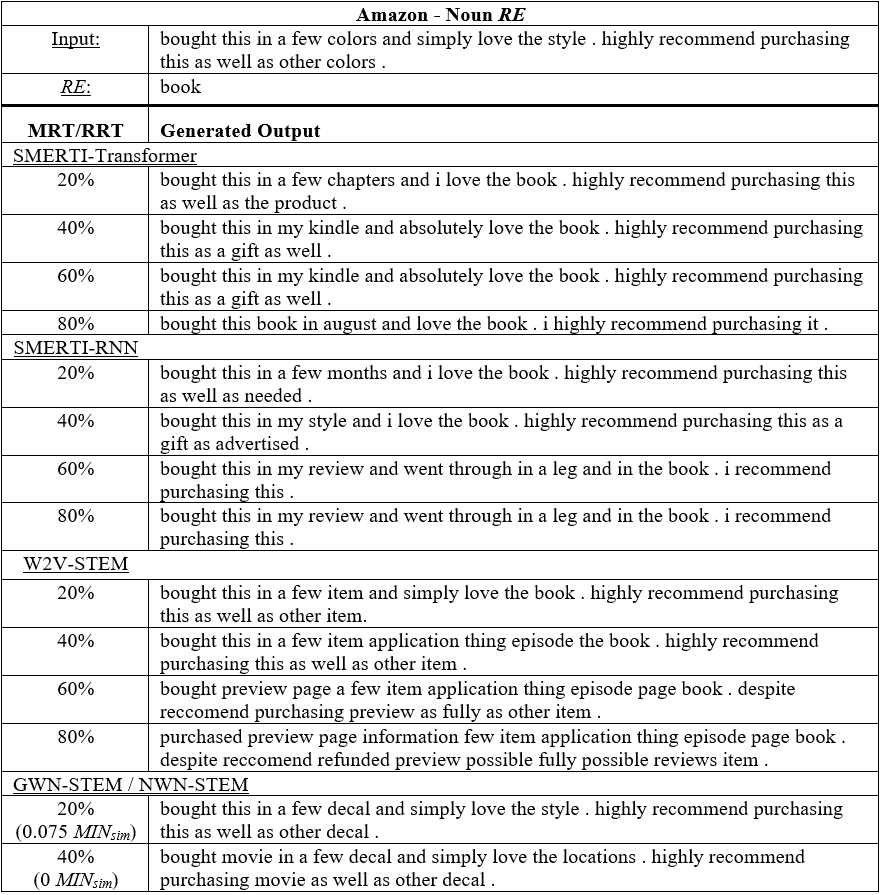}}\\
\end{tabular}
  \caption{\label{tab:amazon_noun} Example outputs for an Amazon evaluation line with noun \RE }
\end{table*}

\begin{table*}
\begin{tabular}{ccc}
\multicolumn{1}{c}{\includegraphics[width=0.95\textwidth]{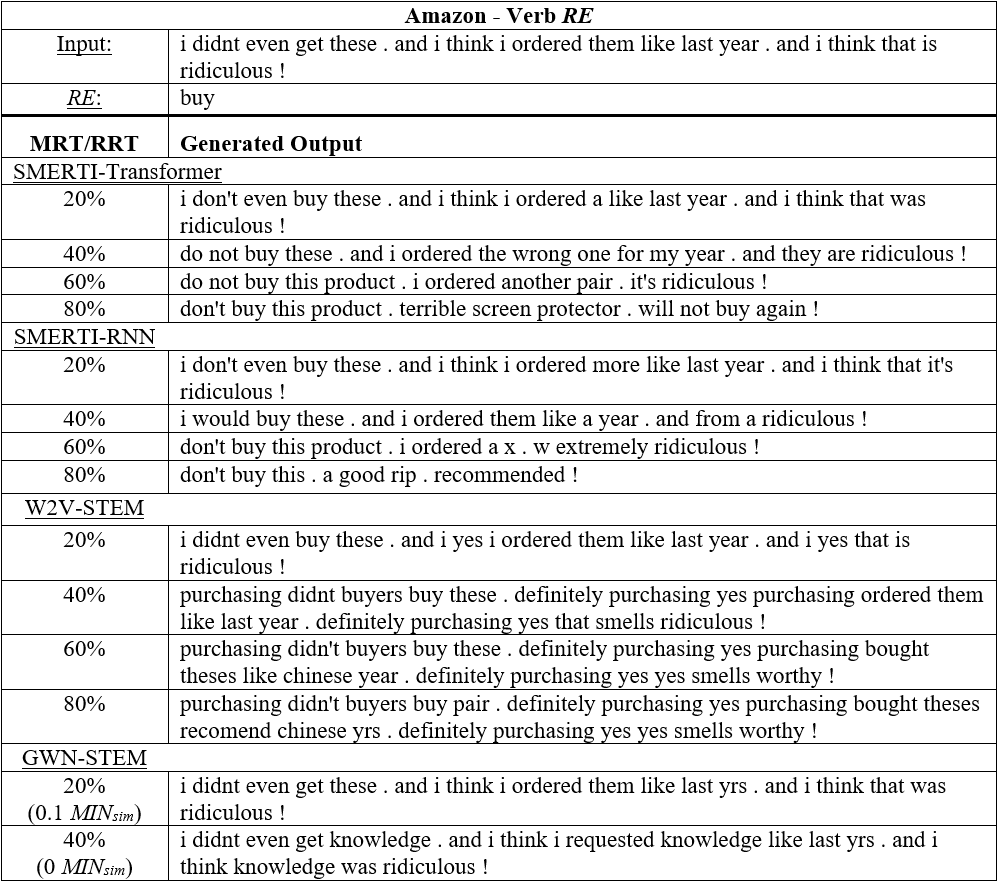}}\\
\end{tabular}
  \caption{\label{tab:amazon_verb} Example outputs for an Amazon evaluation line with verb \RE }
\end{table*}

\begin{table*}
\begin{tabular}{ccc}
\multicolumn{1}{c}{\includegraphics[width=0.95\textwidth]{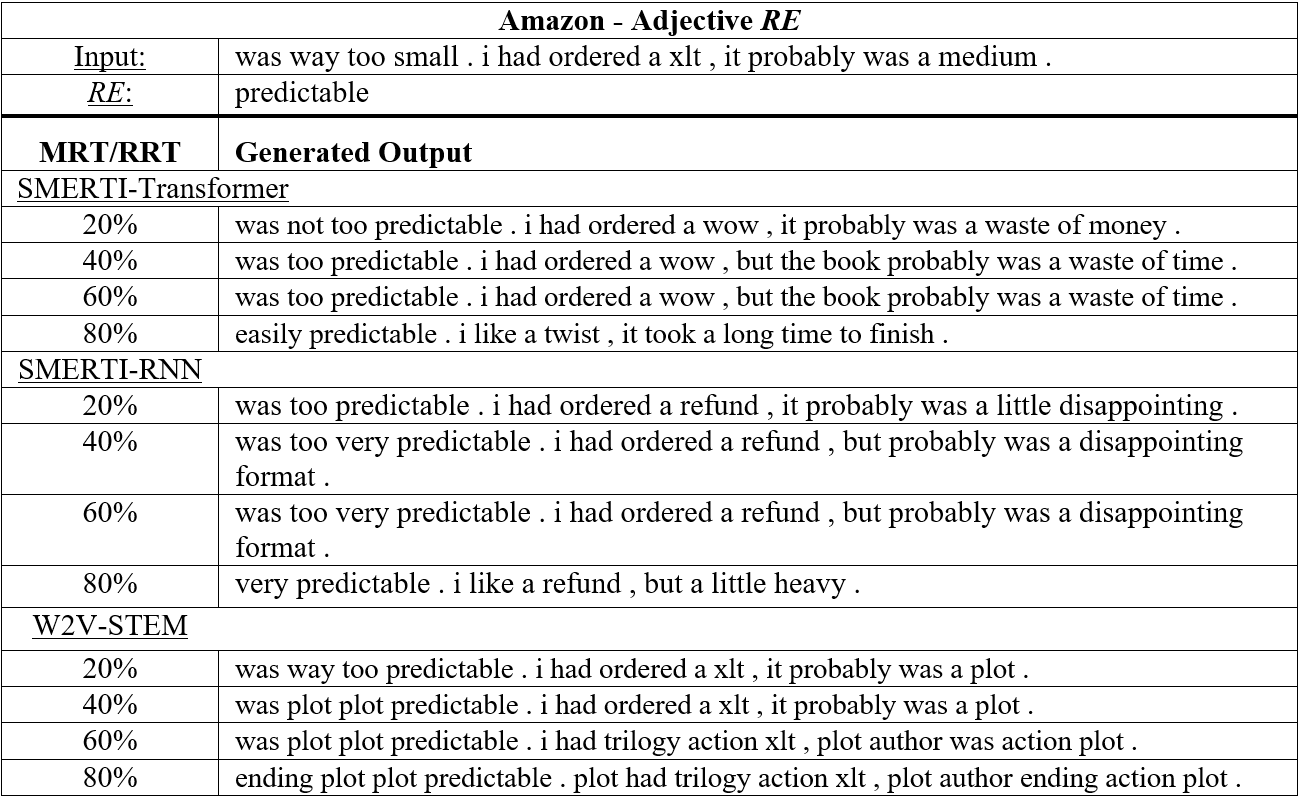}}\\
\end{tabular}
  \caption{\label{tab:amazon_adj} Example outputs for an Amazon evaluation line with adjective \RE }
\end{table*}

\begin{table*}
\begin{tabular}{ccc}
\multicolumn{1}{c}{\includegraphics[width=0.95\textwidth]{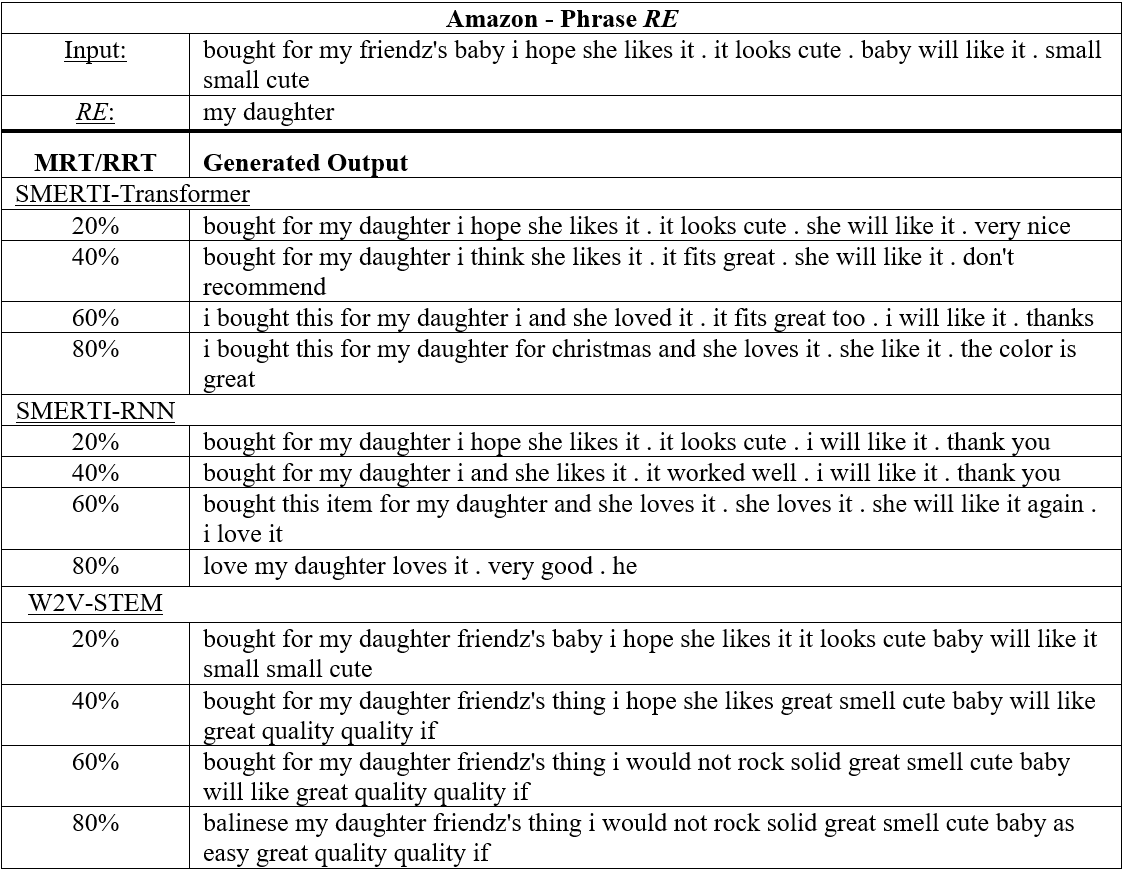}}\\
\end{tabular}
  \caption{\label{tab:amazon_phrase} Example outputs for an Amazon evaluation line with phrase \RE }
\end{table*}

\begin{table*}
\begin{tabular}{ccc}
\multicolumn{1}{c}{\includegraphics[width=0.95\textwidth]{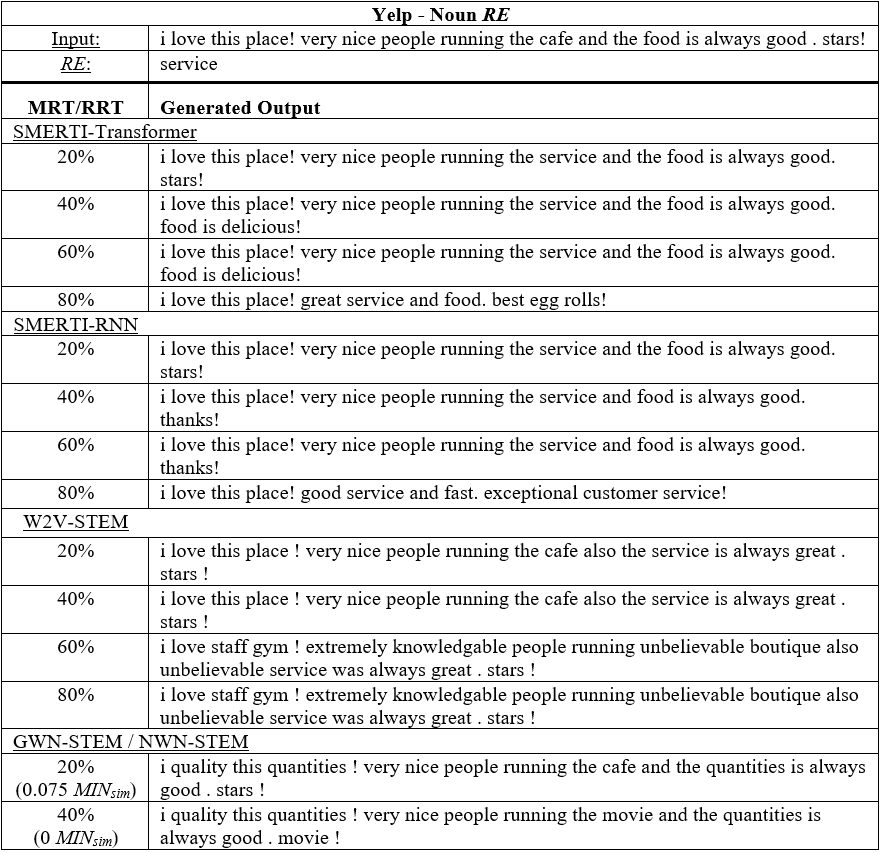}}\\
\end{tabular}
  \caption{\label{tab:yelp_noun} Example outputs for a Yelp evaluation line with noun \RE }
\end{table*}

\begin{table*}
\begin{tabular}{ccc}
\multicolumn{1}{c}{\includegraphics[width=0.95\textwidth]{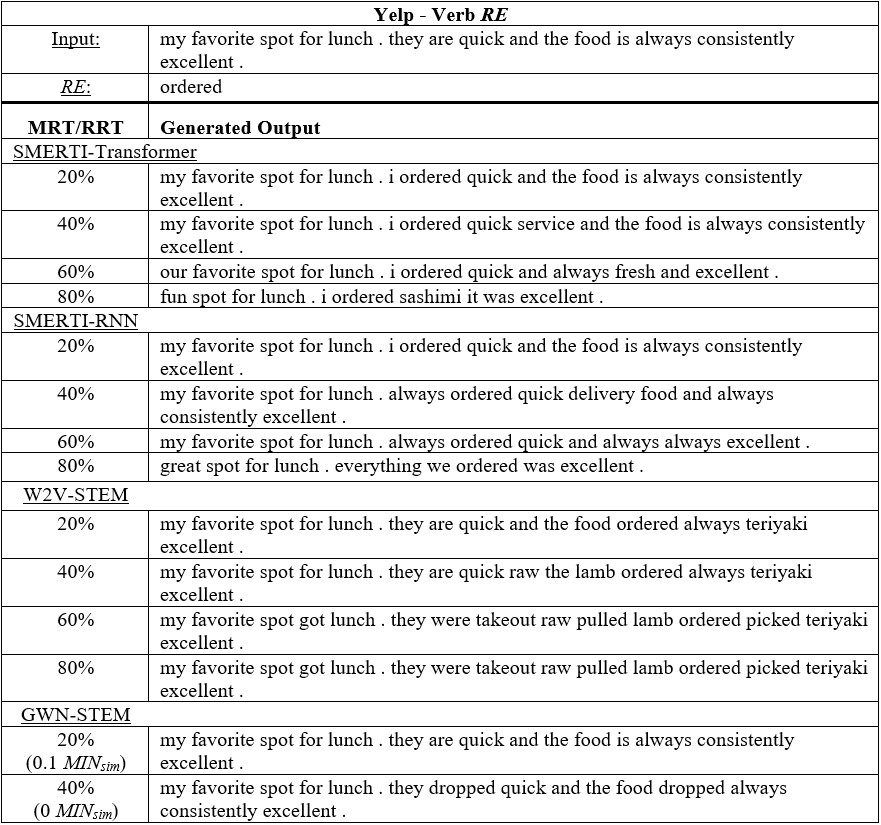}}\\
\end{tabular}
  \caption{\label{tab:yelp_verb} Example outputs for a Yelp evaluation line with verb \RE }
\end{table*}

\begin{table*}
\begin{tabular}{ccc}
\multicolumn{1}{c}{\includegraphics[width=0.95\textwidth]{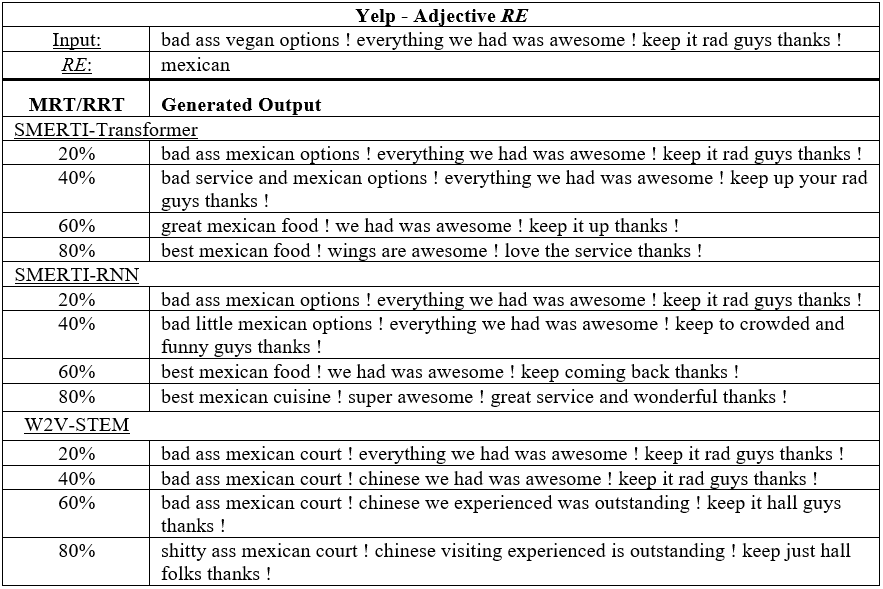}}\\
\end{tabular}
  \caption{\label{tab:yelp_adjective} Example outputs for a Yelp evaluation line with adjective \RE }
\end{table*}

\begin{table*}
\begin{tabular}{ccc}
\multicolumn{1}{c}{\includegraphics[width=0.95\textwidth]{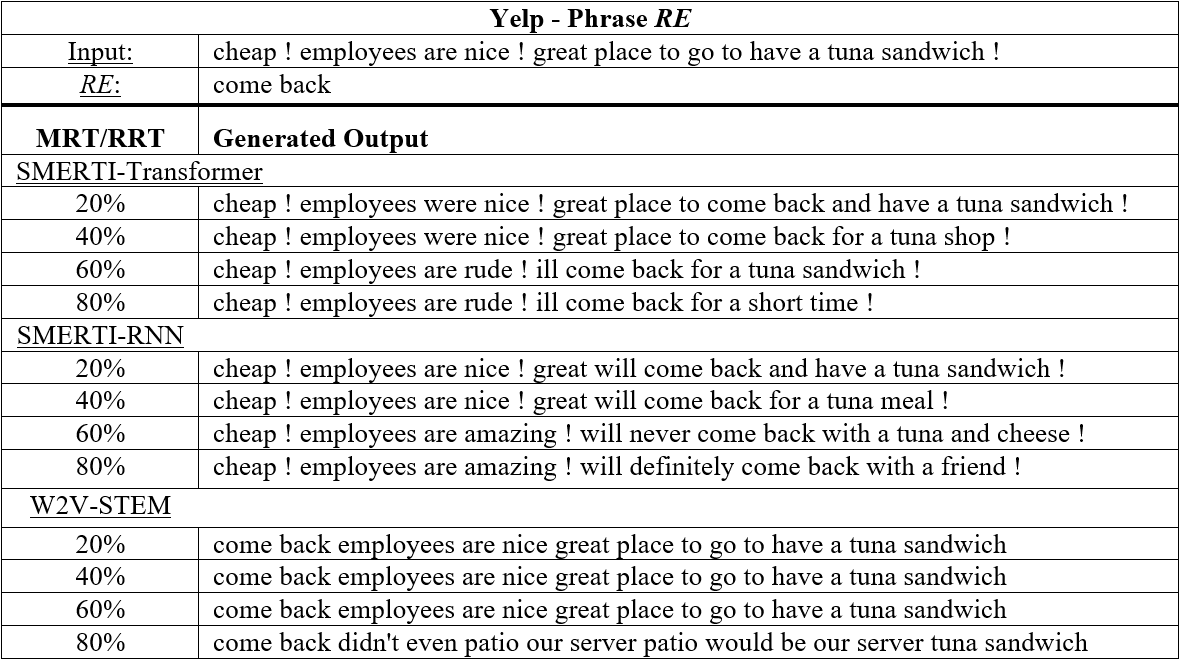}}\\
\end{tabular}
  \caption{\label{tab:yelp_phrase} Example outputs for a Yelp evaluation line with phrase \RE }
\end{table*}

\begin{table*}
\begin{tabular}{ccc}
\multicolumn{1}{c}{\includegraphics[width=0.95\textwidth]{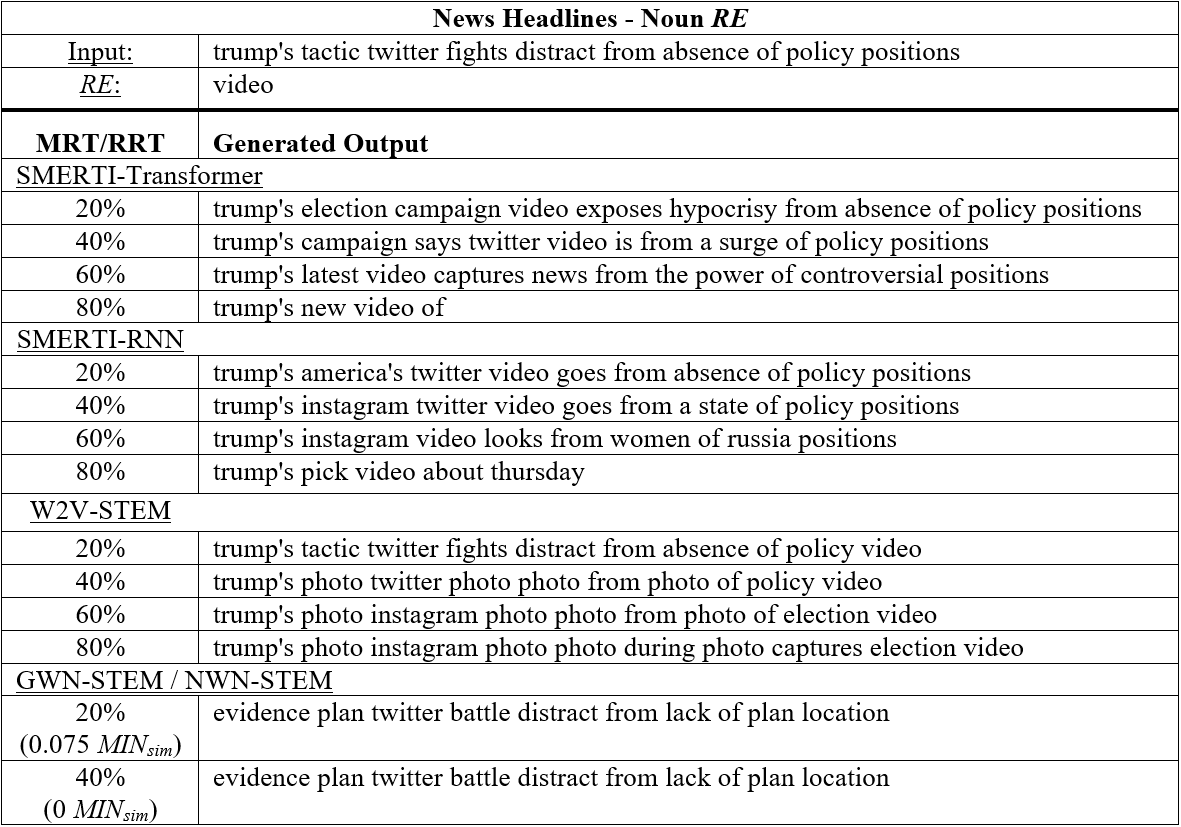}}\\
\end{tabular}
  \caption{\label{tab:news_noun} Example outputs for a news headlines evaluation line with noun \RE }
\end{table*}

\begin{table*}
\begin{tabular}{ccc}
\multicolumn{1}{c}{\includegraphics[width=0.95\textwidth]{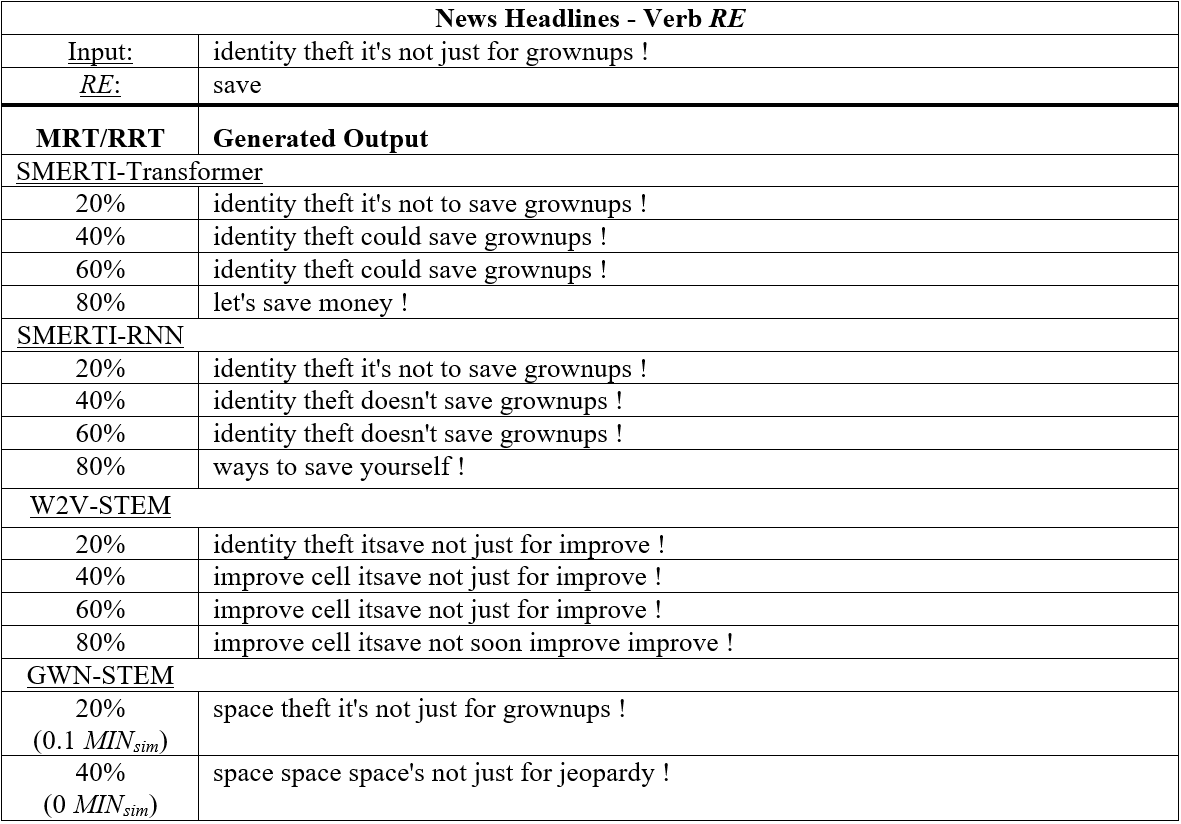}}\\
\end{tabular}
  \caption{\label{tab:news_verb} Example outputs for a news headlines evaluation line with verb \RE }
\end{table*}

\begin{table*}
\begin{tabular}{ccc}
\multicolumn{1}{c}{\includegraphics[width=0.95\textwidth]{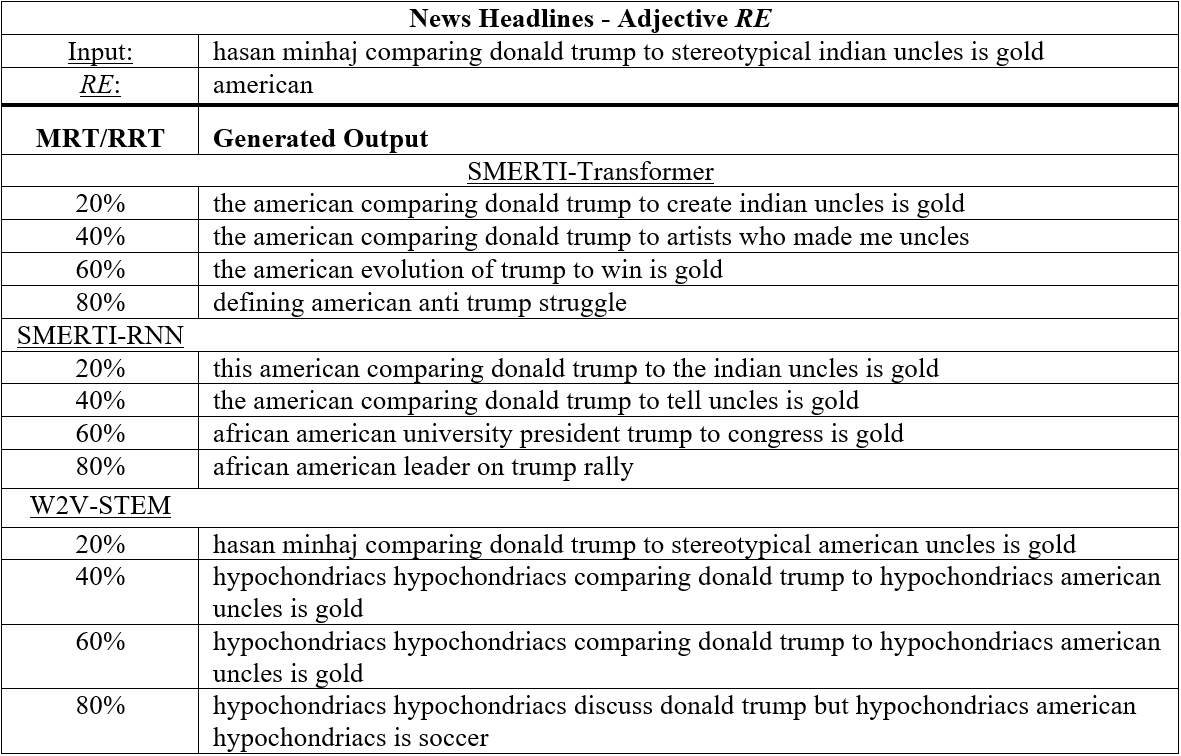}}\\
\end{tabular}
  \caption{\label{tab:news_adj} Example outputs for a news headlines evaluation line with adjective \RE }
\end{table*}

\begin{table*}
\begin{tabular}{ccc}
\multicolumn{1}{c}{\includegraphics[width=0.95\textwidth]{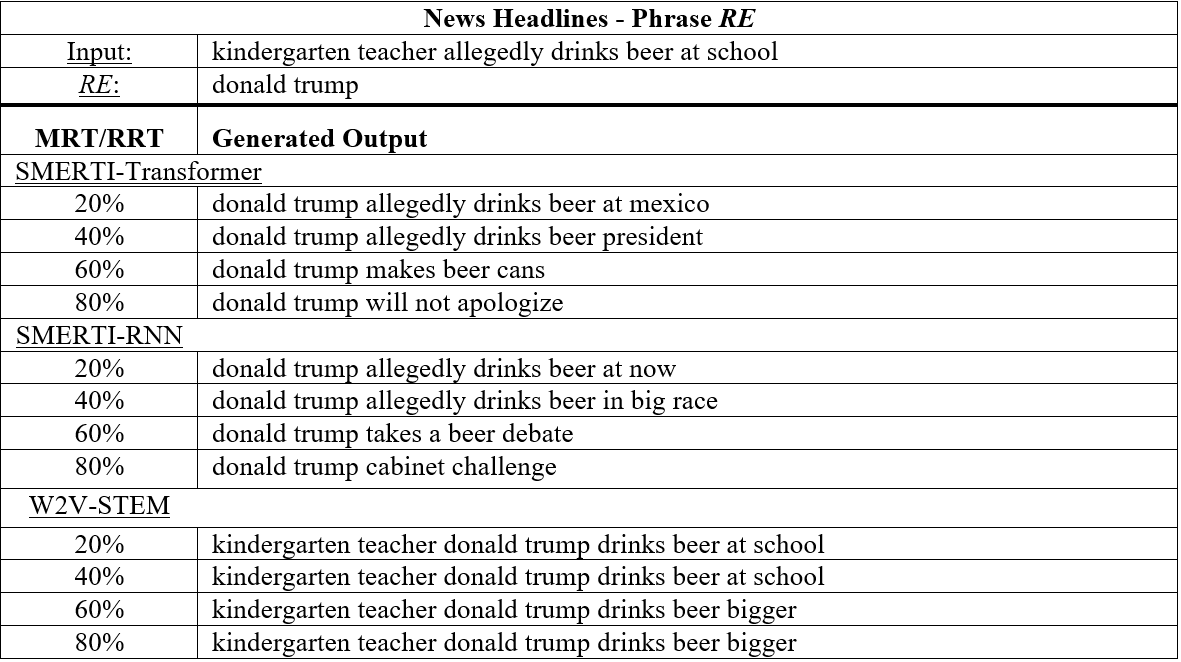}}\\
\end{tabular}
  \caption{\label{tab:news_phrase} Example outputs for a news headlines evaluation line with phrase \RE }
\end{table*}

\bibliography{emnlp-ijcnlp-2019}
\bibliographystyle{acl_natbib}